\documentclass[
]{ceurart}

\usepackage{listings}
\begin{document}

\copyrightyear{2025}
\copyrightclause{Copyright for this paper by its authors.
  Use permitted under Creative Commons License Attribution 4.0
  International (CC BY 4.0).}

\conference{Version as accepted at the BioASQ Lab at CLEF 2025}

\title{Can Language Models Critique Themselves? Investigating Self-Feedback for Retrieval Augmented Generation at BioASQ 2025}

\title[mode=sub]{Notebook for the BioASQ Lab at CLEF 2025}


\author[1]{Samy Ateia}[%
email=Samy.Ateia@stud.uni-regensburg.de,
orcid=0009-0000-2622-9194,
]
\author[1]{Udo Kruschwitz}[%
email=udo.kruschwitz@ur.de,
orcid=0000-0002-5503-0341,
]

\address[1]{Information Science, University of Regensburg, Universitätsstraße 31, 93053, Regensburg, Germany}


\begin{abstract}
  Agentic Retrieval Augmented Generation (RAG) and 'deep research' systems aim to enable autonomous search processes where Large Language Models (LLMs) iteratively refine outputs. However, applying these systems to domain-specific professional search, such as biomedical research, presents challenges, as automated systems may reduce user involvement and misalign with expert information needs. Professional search tasks often demand high levels of user expertise and transparency. The BioASQ CLEF 2025 challenge, using expert-formulated questions, can serve as a platform to study these issues. We explored the performance of current reasoning and nonreasoning LLMs like Gemini-Flash 2.0, o3-mini, o4-mini and DeepSeek-R1. A key aspect of our methodology was a self-feedback mechanism where LLMs generated, evaluated, and then refined their outputs for query expansion and for multiple answer types (yes/no, factoid, list, ideal). We investigated whether this iterative self-correction improves performance and if reasoning models are more capable of generating useful feedback. Preliminary results indicate varied performance for the self-feedback strategy across models and tasks. This work offers insights into LLM self-correction and informs future work on comparing the effectiveness of LLM-generated feedback with direct human expert input in these search systems.
\end{abstract}

\begin{keywords}
  Retrieval Augmented Generation \sep
  Large Language Models \sep
  Biomedical Question Answering \sep
  Professional Search \sep
  Self-Feedback Mechanisms \sep
  Query Expansion \sep
  BioASQ \sep
\end{keywords}

\maketitle

\section{Introduction}
Large Language Models (LLMs) are increasingly deployed in generative search engines that are embedded and offered through AI services such as ChatGPT or Microsoft Copilot. These systems are advertised and used to solve complex, often work-related tasks \cite{suri2024usegenerativesearchengines, edelman2023measuring}. Their newly introduced "deep research"\footnote{\url{https://web.archive.org/web/20250516004701/https://openai.com/index/introducing-deep-research/}}\footnote{\url{https://web.archive.org/web/20250519025641/https://gemini.google/overview/deep-research/}} modes extend generative search by automating iterative, multistep information discovery to produce extensive reports. Tools like Elicit\footnote{\url{https://web.archive.org/web/20250502093323/https://elicit.com/solutions/systematic-reviews}} demonstrate how similar approaches can be tailored specifically for professional search tasks, including technology-assisted systematic literature reviews \cite{bron2024combining}. However, these professional search scenarios typically require significant user expertise and system transparency \cite{verbeneprofsearch, 10.1145/3674127.3674141}. But current LLM limitations, such as the potential for generating unsupported statements \cite{liu2023evaluating}, present challenges, especially when direct expert oversight is reduced through the promise of automation \cite{10.1145/3706598.3714082}.

In this study, we evaluate the performance of current reasoning and non-reasoning LLMs in retrieving relevant biomedical information and answering expert-formulated questions. Specifically, we investigate whether these models can leverage iterative self-feedback mechanisms to enhance their query expansion and response quality.

\subsection{BioASQ Challenge}
The BioASQ challenge provides a long-running platform for evaluating systems on large-scale biomedical semantic indexing and question answering \cite{BioASQ2025overview}. Participants are tasked with retrieving relevant documents and snippets from biomedical literature (PubMed\footnote{\url{https://pubmed.ncbi.nlm.nih.gov/download/\#annual-baseline}}) and generating precise answers to expert-formulated questions, which can be in yes/no, factoid, list, or ideal summary formats. The structured, domain-specific nature of the BioASQ challenge makes it especially suitable for assessing advanced RAG methods for expert information needs.

\subsection{Our Contribution}
Our team has participated in previous iterations of the BioASQ challenge, examining the performance of various commercial and open-source LLMs, the impact of few-shot learning, and the effects of additional context from knowledge bases \cite{ateiakruschwitz, DBLP:conf/clef/AteiaK24}. In this year's challenge (CLEF 2025), we continued our participation across Task A (document and snippet retrieval), Task A+ (Q\&A with own retrieved documents), and Task B (Q\&A with retrieved and gold documents). Our primary investigation centered on the effectiveness of a self-feedback loop implemented with current LLMs, including Gemini-Flash 2.0, o3-mini, o4-mini, and DeepSeek Reasoner, to evaluate if models can improve their own generated query expansions and answers through self-critique.

\section{Related Work}\label{Related}
This work builds upon recent advancements in Large Language Models (LLMs), few-shot and zero-shot learning, Retrieval Augmented Generation (RAG), and their applications to professional search.

\subsection{Large Language Models}
The field of Natural Language Processing (NLP) has been significantly advanced by Large Language Models, mostly based on the transformer architecture \cite{vaswani2017attention}. Early influential models like BERT (Bidirectional Encoder Representations from Transformers) \cite{devlin-etal-2019-bert} demonstrated the power of pre-training on large text corpora. Parallel developments led to autoregressive models such as the GPT (Generative Pre-trained Transformer) series \cite{radford2018improving, brown2020language}. The capabilities of these models were further improved through techniques like Reinforcement Learning from Human Feedback (RLHF), which helps align LLM outputs with human preferences and instructions, making them better at following prompts \cite{ouyang2022training}. 

Recent months have seen the emergence of numerous so-called reasoning models from various developers, including Google's Gemini 2.5\footnote{\url{https://web.archive.org/web/20250518193243/https://blog.google/technology/google-deepmind/gemini-model-thinking-updates-march-2025/}}, OpenAI's o1\footnote{\url{https://web.archive.org/web/20250518101415/https://openai.com/index/learning-to-reason-with-llms/}} to o4-mini model series and models like DeepSeek R1 \cite{deepseekai2025deepseekr1incentivizingreasoningcapability}. These models build on the idea of Chain of Thought (CoT) prompting \cite{wei2022chain} that showed that models perform better when they are prompted to generate additional tokens in their output that mimic reasoning or thinking steps. Fine-tuning models for this reasoning process and therefore enabling variable scaling of test-time compute \cite{snell2024scaling} enabled further advances in model performance on popular benchmarks. Reinforcement learning on math and coding related datasets called Reinforcement Learning with Verifiable Reward (RLVR) \cite{lambert2025tulu3pushingfrontiers} seems to be a current approach to enable models to find useful reasoning strategies.

Our work uses several of these current reasoning and non-reasoning models to compare their performance in a biomedical RAG setting and to see if these reasoning models are better at generating self-feedback.

\subsection{Few and Zero-Shot Learning}
A key characteristic of modern LLMs is their ability to perform tasks with minimal or no task-specific training data, often referred to as In-Context Learning (ICL).
\textbf{Few-shot learning} allows LLMs to learn a new task by conditioning on a few input-output examples provided directly in the prompt. This approach removes the need for extensive, curated training datasets, a concept popularized by models like GPT-3 \cite{brown2020language}.
\textbf{Zero-shot learning} takes this further, enabling LLMs to perform tasks based solely on a natural language description or a direct question, without any preceding examples. 

Our previous work has demonstrated the competitive performance of both zero-shot and few-shot approaches in the BioASQ challenge \cite{ateiakruschwitz, DBLP:conf/clef/AteiaK24}. These techniques are fundamental to the prompting strategies used in our current experiments, forming the basis for initial query/answer generation before any self-feedback loop.

\subsection{Retrieval Augmented Generation (RAG)}

Retrieval Augmented Generation (RAG) combines the generative capabilities of LLMs with information retrieved from external knowledge sources \cite{lewis2020rag}. This approach aims to ground LLM responses in factual data, thereby reducing the likelihood of hallucinations and improving the reliability and verifiability of generated content \cite{shuster-etal-2021-retrieval-augmentation}. A typical RAG pipeline involves a retriever that fetches relevant documents or snippets, and a generator LLM that synthesizes an answer based on the prompt and the retrieved context. The BioASQ challenge itself can be considered an example of a RAG setup in a specialized domain.

The RAG concept is evolving towards more dynamic and autonomous systems, sometimes termed Agentic RAG or 'deep research' systems \cite{openai2025deepresearch}. One of the first of such systems was WebGPT a fine-tuned version of GPT-3 published by a team at OpenAI in 2021 \cite{nakano2022webgptbrowserassistedquestionansweringhuman}. It took OpenAI another 3 years\footnote{\url{https://web.archive.org/web/20250516083609/https://openai.com/index/webgpt/}} to roll out a similar system to their ChatGPT user base \footnote{\url{https://web.archive.org/web/20250511211101/https://openai.com/index/introducing-chatgpt-search/}}. Their newest models, o3 and o4-mini are trained via reinforcement learning to decide autonomously when and how long to search among using other tools \footnote{\url{https://web.archive.org/web/20250514114152/https://openai.com/index/introducing-o3-and-o4-mini/}}. 
These advanced systems may involve LLM-powered agents performing multistep retrieval, reasoning over the retrieved information, and iteratively refining their outputs or search strategies. The deep research modes offered by both OpenAI and Google take these concepts even further and let the models search for over 5 minutes through up to hundreds of websites before synthesizing a multipage report.

Our test of a self-feedback mechanism, where an LLM critiques and revises its own generated queries and answers, is intended to analyze the abilities of off the shelf LLMs on such tasks. In future work, we plan to switch out the LLM generated feedback with feedback from human experts to compare the effectiveness of human and AI guided search processes.

\subsection{Professional Search}
Professional search refers to information seeking conducted in a work-related context, often by specialists who require high precision, control, and the ability to formulate complex queries \cite{verbeneprofsearch, doi:10.1177/02663821211034079}. Domains such as biomedical research demand robust evidence-based answers, making transparency and the ability to trace information back to source documents crucial \cite{higgins2008cochrane}.
LLMs are increasingly being explored for professional search applications, offering potential benefits like advanced query understanding and generation of evidence-based summaries \cite{bron2024combining}. However, challenges such as LLM hallucinations and the need to align with expert workflows remain significant. 

Our previous work, the BioRAGent system, has focused on making LLM-driven RAG accessible and transparent for biomedical question answering, enabling users to review and customize generated boolean queries in the search process \cite{ateia2025bioragent}. This study builds upon this work by exploring the impact of generated critical feedback on query generation and answer generation, which will be compared against human feedback in future work.

\section{Methodology}
\label{Methods}

We evaluated several Large Language Models (LLMs) in the context of the BioASQ CLEF 2025 Challenge, specifically in Task 12 B, which is structured into Phase A (retrieval), Phase A+ (Q\&A based on retrieved snippets), and Phase B (Q\&A based on additional gold-standard snippets).

\subsection{Models}

The models used were grouped into two categories:
\begin{itemize}
 \item \textbf{Non-reasoning models:}
    \begin{itemize}
        \item Gemini Flash 2.0
        \item Gemini 2.5 Flash (used without explicit reasoning mode)
    \end{itemize}

\item \textbf{Reasoning models:}
    \begin{itemize}
        \item o3-mini
        \item o4-mini (introduced mid-challenge and used in later batches)
        \item DeepSeek Reasoner (initially used but replaced due to slow API)
    \end{itemize}
\end{itemize}
\subsection{Task 12 B Experimental Setup}

We participated in all four batches of Task 13 B and submitted several systems under different configurations. Each batch comprised five runs, covering combinations of baseline prompting, feedback-augmented prompting, and few-shot learning (10-shot).

\subsubsection{Phase A: Document and Snippet Retrieval}

Each model configuration involved one of the following strategies:
\begin{itemize}
    \item \textbf{Baseline}: Direct prompt-based query generation without iteration.
    \item \textbf{Feedback (FB)}: Prompt refinement using self-generated feedback.
    \item \textbf{Few-shot}: Prompting the model with 10 examples of successful queries.
\end{itemize}

UR-IW-1 and UR-IW-3 are paired for comparison, both using the same non-reasoning model (Gemini) with and without feedback, respectively. Similarly, UR-IW-2 and UR-IW-4 form a second pair using a reasoning model with and without feedback. While UR-IW-5 is always configured as a non-reasoning few-shot baseline.

The following table summarizes the configurations for Phase A across all four batches:

\begin{table}[h]
\centering
\caption{Overview of Phase A configurations. UR-IW-1/3 compare non-reasoning models with and without feedback, UR-IW-2/4 compare reasoning models. FB = Feedback. UR-IW-5 10-shot baseline.}
\label{tab:phasea-config}
\begin{tabular}{|c|c|c|c|c|c|}
\hline
\textbf{Batch} & UR-IW-1 & UR-IW-2 & UR-IW-3 & UR-IW-4 & UR-IW-5 \\
\hline
1 & Gemini 2.0 & DeepSeek & Gemini 2.0 + FB & DeepSeek + FB & Gemini 2.0 + 10-shot \\
2 & Gemini 2.0 & o3-mini  & Gemini 2.0 + FB & o3-mini + FB  & Gemini 2.0 + 10-shot \\
3 & Gemini 2.0 & o4-mini  & Gemini 2.0 + FB & o4-mini + FB  & Gemini 2.5 + 10-shot \\
4 & Gemini 2.0 & o4-mini  & Gemini 2.0 + FB & o4-mini + FB  & Gemini 2.5 + 10-shot \\
\hline
\end{tabular}
\end{table}

The non-feedback and few-shot approaches were mostly identical to our last years' participation, feedback in phase-A was only used for the query generation and refinement step. The top 10 results from the initial query were passed on to the feedback generating model as additional context. For snippet extraction and reranking no feedback was used.

\subsubsection{Phase A+ and Phase B: Answer Generation}

The system configurations used for Phase A+ and Phase B were similar to those of Phase A. However, the source of contextual snippets to ground the answer generation differed:

\begin{itemize}
    \item \textbf{Phase A+}: Used the top-20 snippets retrieved by the corresponding model in Phase A.
    \item \textbf{Phase B}: Used a merged set combining the top-20 retrieved snippets from Phase A and the gold-standard snippets provided by the organizers.
\end{itemize}

As in Phase A, UR-IW-1/3 and UR-IW-2/4 are grouped to compare feedback vs. non-feedback performance for non-reasoning and reasoning models, respectively. UR-IW-5 serves as a consistent few-shot baseline using non-reasoning models.

\begin{table}[h]
\centering
\caption{Phase A+ and Phase B configurations. Identical setups used across both phases, with differing snippet inputs. UR-IW-1/3 compare non-reasoning models with and without feedback, UR-IW-2/4 compare reasoning models, and UR-IW-5 is a non-reasoning 10-shot baseline.}
\label{tab:phaseaplusb-config}
\begin{tabular}{|c|c|c|c|c|c|}
\hline
\textbf{Batch} & UR-IW-1 & UR-IW-2 & UR-IW-3 & UR-IW-4 & UR-IW-5 \\
\hline
1 & Gemini 2.0 & o3-mini & Gemini 2.0 + FB & o3-mini + FB & Gemini 2.0 + 10-shot \\
2 & Gemini 2.0 & o3-mini & Gemini 2.0 + FB & o3-mini + FB & Gemini 2.0 + 10-shot \\
3 & Gemini 2.0 & o4-mini & Gemini 2.0 + FB & o4-mini + FB & Gemini 2.0 + 10-shot \\
4 & Gemini 2.0 & o4-mini & Gemini 2.0 + FB & o4-mini + FB & Gemini 2.0 + 10-shot \\
\hline
\end{tabular}
\end{table}

The non-feedback and few-shot approaches were again mostly identical to our previous year's participation. For feedback-enhanced runs (UR-IW-3 and UR-IW-4), an additional feedback and refinement step was introduced between draft and final answers. This mechanism relied on task-specific feedback prompts followed by a final revision prompt.

The feedback prompts varied by answer type and were designed to elicit critical evaluation from the model:

\begin{itemize}
    \item \textbf{Yes/No questions}: \textit{``Evaluate the draft answer ('yes' or 'no') against the provided snippets and the question. Indicate explicitly if it should change, with brief reasoning.''}
    \item \textbf{Factoid questions}: \textit{``Evaluate the draft JSON entity list answer against the provided snippets and the question. Clearly suggest corrections, removals, or additions.''}
    \item \textbf{List questions}: Same as factoid prompt.
    \item \textbf{Ideal answer (summary)}: \textit{``Evaluate the provided summary answer for accuracy, clarity, and completeness against the provided snippets and the question. Clearly suggest improvements.''}
\end{itemize}

The generated feedback was then injected into a fixed refinement prompt to guide the model toward a final improved answer:

\begin{quote}
\texttt{Expert Feedback: \{feedback\_response\}}\\
\texttt{Revise and provide the final improved answer strictly following the original instructions.}
\end{quote}

This two-step feedback-refinement process aimed to simulate expert review and enforce more robust quality control over generated answers. 

\subsection{Technical Implementation}

All pipelines were implemented using Python notebooks and the OpenAI, Google and DeepSeek APIs. Query expansion used the \texttt{query\_string} syntax of Elasticsearch. The PubMed annual baseline of 2024 was indexed (title, abstract only) on an Elasticsearch index using the standard English analyzer. Snippet extraction and reranking were performed via LLM prompts. Code and notebooks are publicly available on GitHub\footnote{\url{https://github.com/SamyAteia/bioasq2025}} to ensure full reproducibility.

\section{Results}\label{Results}
As the final results of the BioASQ 2025 Challenge are still being rated by experts and won't be released before September, we can only report on the preliminary results published on the BioASQ website. We participated in Task A (document and snippet retrieval), Task A+ (question answering with own retrieved documents), and Task B (question answering with gold standard documents). The experiments were designed to evaluate the efficacy of different large language models (LLMs) and the impact of self-generated feedback. All results are preliminary and subject to change following manual expert evaluation.

\subsection{Model Selection}
We tested multiple of the current available models with different settings on a small subset of the BioASQ training set \cite{QAcorpusBioASQ} from last year, specifically the fourth batch of BioASQ 12 Task B Phase B.
These models included:

\begin{itemize}
  \item deepseek-reasoner
  \item deepseek-chat
  \item gemini-2.5-pro-exp-02-05
  \item gemini-2.0-flash-thinking-exp-01-21
  \item gemini-2.0-pro-exp-02-05
  \item gemini-2.0-flash-lite
  \item gemini-2.0-flash
  \item claude-3-5-haiku-2024102
  \item claude-3-7-sonnet-20250219
  \item gpt-4.5-preview-2025-02-27
  \item o3-mini-2025-01-31
  \item gpt-4o-mini-2024-07-18
\end{itemize}

\begin{table}[h!]
\centering
\caption{Preliminary LLM Test Results Summary (best in \textbf{bold})}
\label{tab:llm_prelim_summary}
\begin{tabular}{|c|c|c|c|c|}
\hline
Model Name & Yes/No Macro F1 & Factoid MRR & List F-measure & Ideal R2 F1 \\
\hline
gpt-4o-mini-2024-07-18 & 0.911 & 0.544 & 0.531 & 0.308 \\
gpt-4o-mini-self-feedback & 0.833 & 0.544 & 0.530 & 0.307 \\
gpt-4o-mini-o3-feedback & 0.833 & 0.526 & 0.550 & 0.308 \\
o3-mini-2025-01-31 & 0.917 & 0.605 & 0.541 & 0.278 \\
gpt-4.5-preview-2025-02-27 & \textbf{0.957} & 0.675 & 0.546 & 0.398 \\
claude-3-7-sonnet-20250219 & 0.878 & 0.658 & 0.540 & 0.277 \\
claude-3-5-haiku-20241022 & \textbf{0.957} & 0.658 & 0.482 & 0.240 \\
gemini-2.0-flash & 0.954 & \textbf{0.684} & 0.539 & 0.471 \\
gemini-2.0-flash 10-shot & 0.917 & 0.632 & 0.572 & \textbf{0.566} \\
gemini-2.0-flash-lite & 0.911 & 0.632 & \textbf{0.588} & 0.480 \\
gemini-2.0-pro-exp-02-05 & 0.841 & 0.605 & 0.360 & 0.294 \\
gemini-2.0-flash-thinking-exp-01-21 & 0.871 & 0.675 & 0.584 & 0.361 \\
gemini-2.5-pro-exp-02-05 & 0.841 & 0.605 & 0.360 & 0.294 \\
deepseek-chat & 0.862 & 0.596 & 0.559 & 0.329 \\
deepseek-chat-new & 0.911 & 0.632 & 0.544 & 0.282 \\
deepseek-reasoner & \textbf{0.957} & 0.553 & 0.544 & 0.208 \\
\bottomrule
\end{tabular}
\end{table}

Key observations from these preliminary tests include:

gemini-2.0-flash: Demonstrated strong performance across multiple metrics, particularly in yesno\_macro\_f1 (0.954), and factoid\_mrr (0.684), while being competitively priced.

deepseek-reasoner: Achieved high yesno\_accuracy (0.962963) and yesno\_macro\_f1 (0.957075), comparable to gemini-2.0-flash, though with slightly lower performance in factoid and list question types in these preliminary tests.

We decided to choose gemini-2.0-flash as the non-reasoning LLM and also for our 10-shot baseline, as it was both competitive, fast and cheap. For the reasoning model we chose deepseek-reasoner because it is an open-weight model, cheaper to use via the official API and competitive with the other alternative reasoning models (o3-mini, gemini-2.0-flash-thinking).

\subsection{Task A: Document and Snippet Retrieval}
\label{ssec:task_a_results}

In Task A, systems were evaluated on their ability to retrieve relevant documents and snippets for given biomedical questions. Our systems were compared against other participating systems, with the "Top Competitor" representing the leading system in each batch.

Detailed preliminary result tables are available in Appendix \ref{appendixResults}.

\paragraph{Document Retrieval:}
Across the four test batches, our systems demonstrated varied performance.
\begin{itemize}
    \item \textbf{Batch 1}: UR-IW-5 (\texttt{gemini flash 2.0 + 10-shot}) was our top performer, ranking 22nd with a \textbf{MAP} of 0.2865, compared to the Top Competitor's \textbf{MAP} of 0.4246. Our other systems followed, with UR-IW-4 (\texttt{deepseek-reasoner + feedback}) having the lowest \textbf{MAP} (0.1739) among our submissions in this batch.
    \item \textbf{Batch 2}: UR-IW-5 again led our systems (25th, \textbf{MAP} 0.2634), with UR-IW-4 (\texttt{o3-mini + feedback}) closely following (26th, \textbf{MAP} 0.2601). The Top Competitor achieved a \textbf{MAP} of 0.4425.
    \item \textbf{Batch 3}: UR-IW-5 (\texttt{gemini-2.5-flash-preview + 10-shot}) was our best system (24th, \textbf{MAP} 0.1834). The Top Competitor's \textbf{MAP} was 0.3236.
    \item \textbf{Batch 4}: UR-IW-5 (\texttt{gemini-2.5-flash-preview + 10-shot}) ranked 27th with a \textbf{MAP} of 0.0794, while the Top Competitor had a \textbf{MAP} of 0.1801.
\end{itemize}

\paragraph{Snippet Retrieval:}
Similar trends were observed in snippet retrieval performance.
\begin{itemize}
    \item \textbf{Batch 1}: UR-IW-5 (\texttt{gemini flash 2.0 + 10-shot}) performed best among our systems (8th, \textbf{MAP} 0.2768), with the Top Competitor achieving a \textbf{MAP} of 0.4535.
    \item \textbf{Batch 2}: UR-IW-5 again led our entries (12th, \textbf{MAP} 0.3080). The Top Competitor's \textbf{MAP} was 0.5522.
    \item \textbf{Batch 3}: UR-IW-1 (\texttt{gemini flash 2.0}) and UR-IW-5 (\texttt{gemini-2.5-flash-preview + 10-shot}) were our strongest performers, ranking 15th (\textbf{MAP} 0.1534) and 18th (\textbf{MAP} 0.1488) respectively. The Top Competitor had a \textbf{MAP} of 0.4322.
    \item \textbf{Batch 4}: UR-IW-5 (\texttt{gemini-2.5-flash-preview + 10-shot}) was our top system (18th, \textbf{MAP} 0.0511). The Top Competitor achieved a \textbf{MAP} of 0.1634.
\end{itemize}
Generally, the 10-shot run with \texttt{Gemini Flash 2.0 or 2.5} (UR-IW-5) tended to perform better in document and snippet retrieval tasks compared to our other configurations. The impact of feedback on retrieval tasks (UR-IW-3 and UR-IW-4) varied across batches and didn't consistently outperform the base models or the 10-shot variants in \textbf{MAP} scores.

\subsection{Task A+: Question Answering (Own Retrieved Documents)}
\label{ssec:task_aplus_results}

Task A+ required systems to answer questions based on the documents and snippets they retrieved in Phase A.

\paragraph{Yes/No Questions:}
\begin{itemize}
    \item UR-IW-1 (\texttt{gemini flash 2.0}) and UR-IW-2 (\texttt{o3-mini}) often performed strongly. In Batch 1, UR-IW-1, UR-IW-2 and UR-IW-4 (\texttt{o3-mini + feedback}) achieved perfect accuracy and \textbf{Macro F1} scores.
    \item UR-IW-5 (\texttt{gemini flash 2.0 + 10-shot}) achieved a perfect score in Batch 2.
    \item In Batch 4, UR-IW-2 (\texttt{o4-mini}) was our top performer (2nd, \textbf{Macro F1} 0.9097).
    \item The feedback mechanism (UR-IW-3, UR-IW-4) showed mixed results, sometimes improving (e.g., UR-IW-4 in Batch 3) and sometimes underperforming compared to non-feedback versions.
\end{itemize}

\paragraph{Factoid Questions:}
\begin{itemize}
    \item In Batch 1, UR-IW-2 (\texttt{o3-mini}) and UR-IW-5 (\texttt{gemini flash 2.0 + 10-shot}) were our best systems (7th and 8th, \textbf{MRR} 0.3782 and 0.3750 respectively).
    \item UR-IW-4 (\texttt{o3-mini + feedback}) performed well in Batch 2 (2nd, \textbf{MRR} 0.5370).
    \item UR-IW-5 (\texttt{gemini flash 2.0 + 10-shot}) took the top position in Batch 4 with an \textbf{MRR} of 0.5606.
    \item Feedback versions (UR-IW-3, UR-IW-4) had variable performance. For example, UR-IW-3 (\texttt{gemini flash 2.0 + feedback}) ranked 7th in Batch 3 (\textbf{MRR} 0.3100).
\end{itemize}

\paragraph{List Questions:}
\begin{itemize}
    \item Our systems achieved several top rankings in this category. In Batch 1, UR-IW-2 (\texttt{o3-mini}), UR-IW-1 (\texttt{gemini flash 2.0}), UR-IW-5 (\texttt{gemini flash 2.0 + 10-shot}), and UR-IW-4 (\texttt{o3-mini + feedback}) secured the top 4 positions with \textbf{F-Measures} of 0.2567, 0.2411, 0.2395, and 0.2357 respectively.
    \item In Batch 2, UR-IW-2 (\texttt{o3-mini}) was again a strong performer (2nd, \textbf{F-Measure} 0.3805).
    \item The effect of feedback and few-shot learning varied. For instance, in Batch 3, UR-IW-5 (\texttt{gemini flash 2.0 + 10-shot}) ranked 13th (\textbf{F-Measure} 0.3618).
\end{itemize}

\subsection{Task B: Question Answering (Gold Standard Documents)}
\label{ssec:task_b_results}

Task B involved answering questions using additional gold standard documents and snippets.

\paragraph{Yes/No Questions:}
\begin{itemize}
    \item UR-IW-1 (\texttt{gemini flash 2.0}) and UR-IW-5 (\texttt{gemini flash 2.0 + 10-shot}) achieved perfect scores in Batch 1.
    \item UR-IW-5 also achieved a perfect score in Batch 2.
    \item The feedback system UR-IW-4 (\texttt{o3-mini + feedback} or \texttt{o4-mini + feedback}) performed well, often outperforming its non-feedback counterpart in later batches (e.g., UR-IW-4 in Batch 3 and Batch 4 with \textbf{Macro F1} of 0.8706 and 0.9097 respectively).
\end{itemize}

\paragraph{Factoid Questions:}
\begin{itemize}
    \item UR-IW-3 (\texttt{gemini flash 2.0 + feedback}) was our best system in Batch 1 (17th, \textbf{MRR} 0.4821).
    \item In Batch 2, UR-IW-1 (\texttt{gemini flash 2.0}) performed strongly (11th, \textbf{MRR} 0.5926).
    \item UR-IW-4 (\texttt{o4-mini + feedback}) was our top performer in Batch 4 (6th, \textbf{MRR} 0.5909).
    \item The systems with feedback often showed competitive \textbf{MRR} scores, but overall the results were mixed. 
\end{itemize}

\paragraph{List Questions:}
\begin{itemize}
    \item UR-IW-4 (\texttt{o3-mini + feedback} or \texttt{o4-mini + feedback}) consistently performed well, ranking 28th in Batch 1 (\textbf{F-Measure} 0.5069) and 28th in Batch 2 (\textbf{F-Measure} 0.5188).
    \item In Batch 3, UR-IW-5 (\texttt{gemini flash 2.0 + 10-shot}) and UR-IW-3 (\texttt{gemini flash 2.0 + feedback}) were our leading systems.
    \item The results suggest that both few-shot prompting and feedback mechanisms can be beneficial, though their relative effectiveness varied across batches.
\end{itemize}

\section{Discussion and Future Work}\label{Discussion}
\label{ssec:discussion_approaches}

\paragraph{Model Performance:}
Based on the initial model selection tests and the BioASQ task 13B results, \texttt{gemini-2.0-flash} and its variants showed strong and consistent performance, particularly the 10-shot version (UR-IW-5) in retrieval tasks and Yes/No questions. \texttt{o3-mini} and \texttt{o4-mini} (UR-IW-2 and UR-IW-4 configurations) also proved to be competitive, especially in question answering tasks. \texttt{deepseek-reasoner} was competitive, particularly in Task A, Batch 1. But due to the slow API we were unable to complete runs with it in later batches, therefore opting for a proprietary replacement (o3-mini, o4-mini).

\paragraph{Impact of Self-Generated Feedback:}
The motivation to explore self-generated feedback stemmed from our ongoing research into comparing the impact of human expert feedback to LLM generated feedback. In these BioASQ preliminary results, the impact of adding a feedback step (UR-IW-3 and UR-IW-4 configurations) was mixed across all tasks and batches.
For \textbf{Task A (Retrieval)}, feedback configurations did not consistently outperform the base models or 10-shot configurations in terms of \textbf{MAP} scores.
For \textbf{Task A+ and Task B (Question Answering)}, feedback sometimes led to improvements. For instance, in Task B Yes/No questions, UR-IW-4 (with feedback) often surpassed UR-IW-2 (without feedback) in later batches. Similarly, in Task B Factoid questions, feedback systems showed competitive \textbf{MRR} scores.
However, there were also instances where feedback did not lead to better or even resulted in worse performance compared to the base model or the few-shot model. The preliminary tests on model selection also hinted that self-feedback might not always enhance performance for some base models.

\paragraph{Few-Shot Learning vs. Feedback:}
The UR-IW-5 configurations, typically employing \texttt{gemini flash 2.0 + 10-shot} or \texttt{gemini-2.5-flash-preview + 10-shot}, frequently emerged as strong performers, especially in retrieval (Task A) and some question-answering sub-tasks (e.g., Task A+ Factoid Batch 4, Task B Yes/No Batch 1). This suggests that providing a few examples is still a successful way to guide these LLMs.  When comparing \texttt{Gemini Flash 2.0} base (UR-IW-1) with its feedback version (UR-IW-3) and its 10-shot version (UR-IW-5), the 10-shot approach often had an edge, particularly in retrieval.

\paragraph{Best Suited Models and Approaches:}
\begin{itemize}
    \item For \textbf{retrieval tasks (Task A)}, \texttt{gemini flash 2.0 + 10-shot} (UR-IW-5) appeared to be the most promising approach among our submissions.
    \item For \textbf{Yes/No questions (Task A+ \& B)}, \texttt{gemini flash 2.0} (base and 10-shot) and \texttt{o3-mini/o4-mini} (with and without feedback) all showed the ability to achieve high or perfect scores.
    \item For \textbf{Factoid questions (Task A+ \& B)}, performance was more varied. \texttt{o3-mini/o4-mini} with feedback (UR-IW-4) and \texttt{gemini flash 2.0 + 10-shot} (UR-IW-5) had good performances in certain batches.
    \item For \textbf{List questions (Task A+ \& B)}, \texttt{o3-mini} (UR-IW-2) had particularly strong showings in Task A+, Batch 1 and 2. In Task B, \texttt{o3-mini/o4-mini} with feedback (UR-IW-4) also performed well.
\end{itemize}
The choice of "best" model and approach appears to be task-dependent. Few-shot learning with \texttt{gemini-2.0-flash} seems broadly effective. The feedback mechanism shows potential but requires further refinement to ensure consistent improvements across diverse tasks and models. The preliminary test data indicated that \texttt{gemini-2.0-flash} had strong baseline factoid performance, which was reflected in some of the task results.

These are preliminary observations, and a more in-depth analysis will be conducted once the final, manually evaluated results are available. Future work will involve a more granular analysis of the generated answers and the types of errors made by different models and approaches to refine our strategies for future BioASQ challenges. The example code used for feedback and few-shot prompting can be found online\footnote{\url{https://github.com/SamyAteia/bioasq2025}}.

\section{Ethical Considerations}\label{Ethics}
Even if the accuracy and reliability of LLM generated answers in RAG improve, they still tend to make subtle errors or hallucinate information that is not supported by the source documents.
These errors can be especially difficult to catch when expert information needs such as the questions posed in the BioASQ challenge are answered.
The output of these systems should therefore not be used to inform clinical decision-making without thorough expert oversight.

Another ethical issue is the environmental costs of complex multistep RAG systems. As each LLM call is processed on GPU clusters with SOTA models having billions of parameters distributed over these GPUs, every call produces considerably more co2 than a simple TF\_IDF based search result ranking.

\section{Conclusion}\label{Conclusions}
Overall, our feedback-based approach returned mixed results. There was no clear improvement over the zero-shot baselines with the same models. The few-shot approach from last year's participation that we reused as a baseline this year, was, according to the preliminary results, still the most competitive approach from our runs. It was also interesting to see that in our model selection test, the presumably cheaper and smaller distilled models (Gemini flash) were achieving better results than their pricier and presumably bigger counterparts (Gemini Pro) or the reasoning models (o3-mini, DeepSeek R1).

We will build on the introduced feedback approach in future work, comparing the impact of human and LLM generated feedback on overall task performance in professional search \cite{ateia2025professionalsearch}.
We believe this will be a valuable contribution to assess the performance of systems that foster human engagement vs. systems that promise full automation.

\begin{acknowledgments}
We thank the organizers of the BioASQ challenge for their continued support and quick response time.
This work is supported by the German Research Foundation (DFG) as part of the NFDIxCS consortium (Grant number: 501930651).
\end{acknowledgments}

\section*{Declaration on Generative AI}
The authors used the following generative-AI tools while preparing this paper\footnote{https://ceur-ws.org/GenAI/Policy.html}:
\begin{itemize}
\item OpenAI ChatGPT (o3, 4o, 4.5 preview) (May 2025) - drafting content, latex formatting, paraphrase and reword.
\item Google Gemini 2.5 Pro (May 2025) -  drafting content, latex formatting, paraphrase and reword.
\item LanguageTool - spellchecking, paraphrase and reword.
\end{itemize}

All AI-generated material was critically reviewed, revised and verified by the human authors. The authors accept full responsibility for the integrity and accuracy of the final manuscript.
\bibliography{bibliography}

\begin{thebibliography}{28}
\expandafter\ifx\csname natexlab\endcsname\relax\def\natexlab#1{#1}\fi
\providecommand{\url}[1]{\texttt{#1}}
\providecommand{\href}[2]{#2}
\providecommand{\path}[1]{#1}
\providecommand{\DOIprefix}{doi:}
\providecommand{\ArXivprefix}{arXiv:}
\providecommand{\URLprefix}{URL: }
\providecommand{\Pubmedprefix}{pmid:}
\providecommand{\doi}[1]{\href{http://dx.doi.org/#1}{\path{#1}}}
\providecommand{\Pubmed}[1]{\href{pmid:#1}{\path{#1}}}
\providecommand{\bibinfo}[2]{#2}
\ifx\xfnm\relax \def\xfnm[#1]{\unskip,\space#1}\fi
\bibitem[{Suri et~al.(2024)Suri, Counts, Wang, Chen, Wan, Safavi, Neville, Shah, White, Andersen, Buscher, Manivannan, Rangan, and Yang}]{suri2024usegenerativesearchengines}
\bibinfo{author}{S.~Suri}, \bibinfo{author}{S.~Counts}, \bibinfo{author}{L.~Wang}, \bibinfo{author}{C.~Chen}, \bibinfo{author}{M.~Wan}, \bibinfo{author}{T.~Safavi}, \bibinfo{author}{J.~Neville}, \bibinfo{author}{C.~Shah}, \bibinfo{author}{R.~W. White}, \bibinfo{author}{R.~Andersen}, \bibinfo{author}{G.~Buscher}, \bibinfo{author}{S.~Manivannan}, \bibinfo{author}{N.~Rangan}, \bibinfo{author}{L.~Yang}, \bibinfo{title}{{The Use of Generative Search Engines for Knowledge Work and Complex Tasks}}, \bibinfo{year}{2024}. \URLprefix \url{https://arxiv.org/abs/2404.04268}. \href{http://arxiv.org/abs/2404.04268}{{\tt arXiv:2404.04268}}.
\bibitem[{Edelman et~al.(2023)Edelman, Ngwe, and Peng}]{edelman2023measuring}
\bibinfo{author}{B.~G. Edelman}, \bibinfo{author}{D.~Ngwe}, \bibinfo{author}{S.~Peng},
\newblock \bibinfo{title}{{Measuring the impact of AI on information worker productivity}},
\newblock \bibinfo{journal}{Available at SSRN 4648686}  (\bibinfo{year}{2023}).
\bibitem[{Bron et~al.(2024)Bron, Greijn, Coimbra, van~de Schoot, and Bagheri}]{bron2024combining}
\bibinfo{author}{M.~P. Bron}, \bibinfo{author}{B.~Greijn}, \bibinfo{author}{B.~M. Coimbra}, \bibinfo{author}{R.~van~de Schoot}, \bibinfo{author}{A.~Bagheri},
\newblock \bibinfo{title}{Combining large language model classifications and active learning for improved technology-assisted review},
\newblock in: \bibinfo{booktitle}{Proceedings of the International Workshop on Interactive Adaptive Learning (IAL@PKDD/ECML 2024)}, volume \bibinfo{volume}{3770} of \textit{\bibinfo{series}{CEUR Workshop Proceedings}}, \bibinfo{publisher}{CEUR-WS.org}, \bibinfo{year}{2024}, pp. \bibinfo{pages}{77--95}. \URLprefix \url{https://ceur-ws.org/Vol-3770/paper6.pdf}.
\bibitem[{Verberne et~al.(2019)Verberne, He, Kruschwitz, Wiggers, Larsen, Russell-Rose, and de~Vries}]{verbeneprofsearch}
\bibinfo{author}{S.~Verberne}, \bibinfo{author}{J.~He}, \bibinfo{author}{U.~Kruschwitz}, \bibinfo{author}{G.~Wiggers}, \bibinfo{author}{B.~Larsen}, \bibinfo{author}{T.~Russell-Rose}, \bibinfo{author}{A.~P. de~Vries},
\newblock \bibinfo{title}{{First International Workshop on Professional Search}},
\newblock \bibinfo{journal}{SIGIR Forum} \bibinfo{volume}{52} (\bibinfo{year}{2019}) \bibinfo{pages}{153–162}. \URLprefix \url{https://doi.org/10.1145/3308774.3308799}. \DOIprefix\doi{10.1145/3308774.3308799}.
\bibitem[{Verberne(2024)}]{10.1145/3674127.3674141}
\bibinfo{author}{S.~Verberne}, \bibinfo{title}{Professional Search}, \bibinfo{edition}{1} ed., \bibinfo{publisher}{Association for Computing Machinery}, \bibinfo{address}{New York, NY, USA}, \bibinfo{year}{2024}, p. \bibinfo{pages}{501–514}. \URLprefix \url{https://doi.org/10.1145/3674127.3674141}.
\bibitem[{Liu et~al.(2023)Liu, Zhang, and Liang}]{liu2023evaluating}
\bibinfo{author}{N.~F. Liu}, \bibinfo{author}{T.~Zhang}, \bibinfo{author}{P.~Liang},
\newblock \bibinfo{title}{{Evaluating verifiability in generative search engines}},
\newblock \bibinfo{journal}{arXiv preprint arXiv:2304.09848}  (\bibinfo{year}{2023}).
\bibitem[{Spatharioti et~al.(2025)Spatharioti, Rothschild, Goldstein, and Hofman}]{10.1145/3706598.3714082}
\bibinfo{author}{S.~E. Spatharioti}, \bibinfo{author}{D.~Rothschild}, \bibinfo{author}{D.~G. Goldstein}, \bibinfo{author}{J.~M. Hofman},
\newblock \bibinfo{title}{{Effects of LLM-based Search on Decision Making: Speed, Accuracy, and Overreliance}},
\newblock in: \bibinfo{booktitle}{Proceedings of the 2025 CHI Conference on Human Factors in Computing Systems}, CHI '25, \bibinfo{publisher}{Association for Computing Machinery}, \bibinfo{address}{New York, NY, USA}, \bibinfo{year}{2025}. \URLprefix \url{https://doi.org/10.1145/3706598.3714082}. \DOIprefix\doi{10.1145/3706598.3714082}.
\bibitem[{Nentidis et~al.(2025)Nentidis, Katsimpras, Krithara, Krallinger, Rodríguez-Ortega, Rodriguez-López, Loukachevitch, Sakhovskiy, Tutubalina, Dimitriadis, Tsoumakas, Giannakoulas, Bekiaridou, Samaras, Maria Di~Nunzio, Ferro, Marchesin, Martinelli, Silvello, and Paliouras}]{BioASQ2025overview}
\bibinfo{author}{A.~Nentidis}, \bibinfo{author}{G.~Katsimpras}, \bibinfo{author}{A.~Krithara}, \bibinfo{author}{M.~Krallinger}, \bibinfo{author}{M.~Rodríguez-Ortega}, \bibinfo{author}{E.~Rodriguez-López}, \bibinfo{author}{N.~Loukachevitch}, \bibinfo{author}{A.~Sakhovskiy}, \bibinfo{author}{E.~Tutubalina}, \bibinfo{author}{D.~Dimitriadis}, \bibinfo{author}{G.~Tsoumakas}, \bibinfo{author}{G.~Giannakoulas}, \bibinfo{author}{A.~Bekiaridou}, \bibinfo{author}{A.~Samaras}, \bibinfo{author}{G.~Maria Di~Nunzio}, \bibinfo{author}{N.~Ferro}, \bibinfo{author}{S.~Marchesin}, \bibinfo{author}{M.~Martinelli}, \bibinfo{author}{G.~Silvello}, \bibinfo{author}{G.~Paliouras},
\newblock \bibinfo{title}{{Overview of BioASQ 2025: The thirteenth BioASQ challenge on large-scale biomedical semantic indexing and question answering}},
\newblock in: \bibinfo{editor}{J.~C. de~Albornoz}, \bibinfo{editor}{J.~Gonzalo}, \bibinfo{editor}{L.~Plaza}, \bibinfo{editor}{A.~G.~S. de~Herrera}, \bibinfo{editor}{J.~Mothe}, \bibinfo{editor}{F.~Piroi}, \bibinfo{editor}{P.~Rosso}, \bibinfo{editor}{D.~Spina}, \bibinfo{editor}{G.~Faggioli}, \bibinfo{editor}{N.~Ferro} (Eds.), \bibinfo{booktitle}{Experimental IR Meets Multilinguality, Multimodality, and Interaction. Proceedings of the Sixteenth International Conference of the CLEF Association (CLEF 2025)}, \bibinfo{year}{2025}.
\bibitem[{Ateia and Kruschwitz(2023)}]{ateiakruschwitz}
\bibinfo{author}{S.~Ateia}, \bibinfo{author}{U.~Kruschwitz},
\newblock \bibinfo{title}{Is chatgpt a biomedical expert?},
\newblock in: \bibinfo{editor}{M.~Aliannejadi}, \bibinfo{editor}{G.~Faggioli}, \bibinfo{editor}{N.~Ferro}, \bibinfo{editor}{M.~Vlachos} (Eds.), \bibinfo{booktitle}{{Working Notes of the Conference and Labs of the Evaluation Forum {(CLEF} 2023), Thessaloniki, Greece, September 18th to 21st, 2023}}, volume \bibinfo{volume}{3497} of \textit{\bibinfo{series}{{CEUR} Workshop Proceedings}}, \bibinfo{publisher}{CEUR-WS.org}, \bibinfo{year}{2023}, pp. \bibinfo{pages}{73--90}. \URLprefix \url{https://ceur-ws.org/Vol-3497/paper-006.pdf}.
\bibitem[{Ateia and Kruschwitz(2024)}]{DBLP:conf/clef/AteiaK24}
\bibinfo{author}{S.~Ateia}, \bibinfo{author}{U.~Kruschwitz},
\newblock \bibinfo{title}{Can open-source llms compete with commercial models? exploring the few-shot performance of current {GPT} models in biomedical tasks},
\newblock in: \bibinfo{editor}{G.~Faggioli}, \bibinfo{editor}{N.~Ferro}, \bibinfo{editor}{P.~Galusc{\'{a}}kov{\'{a}}}, \bibinfo{editor}{A.~G.~S. de~Herrera} (Eds.), \bibinfo{booktitle}{Working Notes of the Conference and Labs of the Evaluation Forum {(CLEF} 2024), Grenoble, France, 9-12 September, 2024}, volume \bibinfo{volume}{3740} of \textit{\bibinfo{series}{{CEUR} Workshop Proceedings}}, \bibinfo{publisher}{CEUR-WS.org}, \bibinfo{year}{2024}, pp. \bibinfo{pages}{78--98}. \URLprefix \url{https://ceur-ws.org/Vol-3740/paper-07.pdf}.
\bibitem[{Vaswani et~al.(2017)Vaswani, Shazeer, Parmar, Uszkoreit, Jones, Gomez, Kaiser, and Polosukhin}]{vaswani2017attention}
\bibinfo{author}{A.~Vaswani}, \bibinfo{author}{N.~Shazeer}, \bibinfo{author}{N.~Parmar}, \bibinfo{author}{J.~Uszkoreit}, \bibinfo{author}{L.~Jones}, \bibinfo{author}{A.~N. Gomez}, \bibinfo{author}{L.~Kaiser}, \bibinfo{author}{I.~Polosukhin},
\newblock \bibinfo{title}{{Attention is All You Need}},
\newblock in: \bibinfo{booktitle}{Proceedings of the 31st International Conference on Neural Information Processing Systems}, NIPS'17, \bibinfo{publisher}{Curran Associates Inc.}, \bibinfo{address}{Red Hook, NY, USA}, \bibinfo{year}{2017}, p. \bibinfo{pages}{6000–6010}.
\bibitem[{Devlin et~al.(2019)Devlin, Chang, Lee, and Toutanova}]{devlin-etal-2019-bert}
\bibinfo{author}{J.~Devlin}, \bibinfo{author}{M.-W. Chang}, \bibinfo{author}{K.~Lee}, \bibinfo{author}{K.~Toutanova},
\newblock \bibinfo{title}{{BERT}: Pre-training of deep bidirectional transformers for language understanding},
\newblock in: \bibinfo{editor}{J.~Burstein}, \bibinfo{editor}{C.~Doran}, \bibinfo{editor}{T.~Solorio} (Eds.), \bibinfo{booktitle}{Proceedings of the 2019 Conference of the North {A}merican Chapter of the Association for Computational Linguistics: Human Language Technologies, Volume 1 (Long and Short Papers)}, \bibinfo{publisher}{Association for Computational Linguistics}, \bibinfo{address}{Minneapolis, Minnesota}, \bibinfo{year}{2019}, pp. \bibinfo{pages}{4171--4186}. \URLprefix \url{https://aclanthology.org/N19-1423/}. \DOIprefix\doi{10.18653/v1/N19-1423}.
\bibitem[{Radford et~al.(2018)Radford, Narasimhan, Salimans, Sutskever et~al.}]{radford2018improving}
\bibinfo{author}{A.~Radford}, \bibinfo{author}{K.~Narasimhan}, \bibinfo{author}{T.~Salimans}, \bibinfo{author}{I.~Sutskever}, et~al., \bibinfo{title}{{Improving language understanding by generative pre-training}}, \bibinfo{howpublished}{preprint}, \bibinfo{year}{2018}. \URLprefix \url{https://web.archive.org/web/20240522131718/https://cdn.openai.com/research-covers/language-unsupervised/language_understanding_paper.pdf}.
\bibitem[{Brown et~al.(2020)Brown, Mann, Ryder, Subbiah, Kaplan, Dhariwal, Neelakantan, Shyam, Sastry, Askell et~al.}]{brown2020language}
\bibinfo{author}{T.~Brown}, \bibinfo{author}{B.~Mann}, \bibinfo{author}{N.~Ryder}, \bibinfo{author}{M.~Subbiah}, \bibinfo{author}{J.~D. Kaplan}, \bibinfo{author}{P.~Dhariwal}, \bibinfo{author}{A.~Neelakantan}, \bibinfo{author}{P.~Shyam}, \bibinfo{author}{G.~Sastry}, \bibinfo{author}{A.~Askell}, et~al.,
\newblock \bibinfo{title}{{Language models are few-shot learners}},
\newblock \bibinfo{journal}{Advances in neural information processing systems} \bibinfo{volume}{33} (\bibinfo{year}{2020}) \bibinfo{pages}{1877--1901}.
\bibitem[{Ouyang et~al.(2022)Ouyang, Wu, Jiang, Almeida, Wainwright, Mishkin, Zhang, Agarwal, Slama, Ray et~al.}]{ouyang2022training}
\bibinfo{author}{L.~Ouyang}, \bibinfo{author}{J.~Wu}, \bibinfo{author}{X.~Jiang}, \bibinfo{author}{D.~Almeida}, \bibinfo{author}{C.~Wainwright}, \bibinfo{author}{P.~Mishkin}, \bibinfo{author}{C.~Zhang}, \bibinfo{author}{S.~Agarwal}, \bibinfo{author}{K.~Slama}, \bibinfo{author}{A.~Ray}, et~al.,
\newblock \bibinfo{title}{{Training language models to follow instructions with human feedback}},
\newblock \bibinfo{journal}{Advances in Neural Information Processing Systems} \bibinfo{volume}{35} (\bibinfo{year}{2022}) \bibinfo{pages}{27730--27744}.
\bibitem[{DeepSeek-AI et~al.(2025)DeepSeek-AI, Guo, Yang, Zhang, Song, Zhang, Xu, Zhu, Ma, Wang, Bi, Zhang, Yu, Wu, Wu, Gou, Shao, Li, Gao, Liu, Xue, Wang, Wu, Feng, Lu, Zhao, Deng, Zhang, Ruan, Dai, Chen, Ji, Li, Lin, Dai, Luo, Hao, Chen, Li, Zhang, Bao, Xu, Wang, Ding, Xin, Gao, Qu, Li, Guo, Li, Wang, Chen, Yuan, Qiu, Li, Cai, Ni, Liang, Chen, Dong, Hu, Gao, Guan, Huang, Yu, Wang, Zhang, Zhao, Wang, Zhang, Xu, Xia, Zhang, Zhang, Tang, Li, Wang, Li, Tian, Huang, Zhang, Wang, Chen, Du, Ge, Zhang, Pan, Wang, Chen, Jin, Chen, Lu, Zhou, Chen, Ye, Wang, Yu, Zhou, Pan, Li, Zhou, Wu, Ye, Yun, Pei, Sun, Wang, Zeng, Zhao, Liu, Liang, Gao, Yu, Zhang, Xiao, An, Liu, Wang, Chen, Nie, Cheng, Liu, Xie, Liu, Yang, Li, Su, Lin, Li, Jin, Shen, Chen, Sun, Wang, Song, Zhou, Wang, Shan, Li, Wang, Wei, Zhang, Xu, Li, Zhao, Sun, Wang, Yu, Zhang, Shi, Xiong, He, Piao, Wang, Tan, Ma, Liu, Guo, Ou, Wang, Gong, Zou, He, Xiong, Luo, You, Liu, Zhou, Zhu, Xu, Huang, Li, Zheng, Zhu, Ma, Tang, Zha, Yan, Ren, Ren, Sha, Fu, Xu, Xie, Zhang,
  Hao, Ma, Yan, Wu, Gu, Zhu, Liu, Li, Xie, Song, Pan, Huang, Xu, Zhang, and Zhang}]{deepseekai2025deepseekr1incentivizingreasoningcapability}
\bibinfo{author}{DeepSeek-AI}, \bibinfo{author}{D.~Guo}, \bibinfo{author}{D.~Yang}, \bibinfo{author}{H.~Zhang}, \bibinfo{author}{J.~Song}, \bibinfo{author}{R.~Zhang}, \bibinfo{author}{R.~Xu}, \bibinfo{author}{Q.~Zhu}, \bibinfo{author}{S.~Ma}, \bibinfo{author}{P.~Wang}, \bibinfo{author}{X.~Bi}, \bibinfo{author}{X.~Zhang}, \bibinfo{author}{X.~Yu}, \bibinfo{author}{Y.~Wu}, \bibinfo{author}{Z.~F. Wu}, \bibinfo{author}{Z.~Gou}, \bibinfo{author}{Z.~Shao}, \bibinfo{author}{Z.~Li}, \bibinfo{author}{Z.~Gao}, \bibinfo{author}{A.~Liu}, \bibinfo{author}{B.~Xue}, \bibinfo{author}{B.~Wang}, \bibinfo{author}{B.~Wu}, \bibinfo{author}{B.~Feng}, \bibinfo{author}{C.~Lu}, \bibinfo{author}{C.~Zhao}, \bibinfo{author}{C.~Deng}, \bibinfo{author}{C.~Zhang}, \bibinfo{author}{C.~Ruan}, \bibinfo{author}{D.~Dai}, \bibinfo{author}{D.~Chen}, \bibinfo{author}{D.~Ji}, \bibinfo{author}{E.~Li}, \bibinfo{author}{F.~Lin}, \bibinfo{author}{F.~Dai}, \bibinfo{author}{F.~Luo}, \bibinfo{author}{G.~Hao}, \bibinfo{author}{G.~Chen},
  \bibinfo{author}{G.~Li}, \bibinfo{author}{H.~Zhang}, \bibinfo{author}{H.~Bao}, \bibinfo{author}{H.~Xu}, \bibinfo{author}{H.~Wang}, \bibinfo{author}{H.~Ding}, \bibinfo{author}{H.~Xin}, \bibinfo{author}{H.~Gao}, \bibinfo{author}{H.~Qu}, \bibinfo{author}{H.~Li}, \bibinfo{author}{J.~Guo}, \bibinfo{author}{J.~Li}, \bibinfo{author}{J.~Wang}, \bibinfo{author}{J.~Chen}, \bibinfo{author}{J.~Yuan}, \bibinfo{author}{J.~Qiu}, \bibinfo{author}{J.~Li}, \bibinfo{author}{J.~L. Cai}, \bibinfo{author}{J.~Ni}, \bibinfo{author}{J.~Liang}, \bibinfo{author}{J.~Chen}, \bibinfo{author}{K.~Dong}, \bibinfo{author}{K.~Hu}, \bibinfo{author}{K.~Gao}, \bibinfo{author}{K.~Guan}, \bibinfo{author}{K.~Huang}, \bibinfo{author}{K.~Yu}, \bibinfo{author}{L.~Wang}, \bibinfo{author}{L.~Zhang}, \bibinfo{author}{L.~Zhao}, \bibinfo{author}{L.~Wang}, \bibinfo{author}{L.~Zhang}, \bibinfo{author}{L.~Xu}, \bibinfo{author}{L.~Xia}, \bibinfo{author}{M.~Zhang}, \bibinfo{author}{M.~Zhang}, \bibinfo{author}{M.~Tang}, \bibinfo{author}{M.~Li},
  \bibinfo{author}{M.~Wang}, \bibinfo{author}{M.~Li}, \bibinfo{author}{N.~Tian}, \bibinfo{author}{P.~Huang}, \bibinfo{author}{P.~Zhang}, \bibinfo{author}{Q.~Wang}, \bibinfo{author}{Q.~Chen}, \bibinfo{author}{Q.~Du}, \bibinfo{author}{R.~Ge}, \bibinfo{author}{R.~Zhang}, \bibinfo{author}{R.~Pan}, \bibinfo{author}{R.~Wang}, \bibinfo{author}{R.~J. Chen}, \bibinfo{author}{R.~L. Jin}, \bibinfo{author}{R.~Chen}, \bibinfo{author}{S.~Lu}, \bibinfo{author}{S.~Zhou}, \bibinfo{author}{S.~Chen}, \bibinfo{author}{S.~Ye}, \bibinfo{author}{S.~Wang}, \bibinfo{author}{S.~Yu}, \bibinfo{author}{S.~Zhou}, \bibinfo{author}{S.~Pan}, \bibinfo{author}{S.~S. Li}, \bibinfo{author}{S.~Zhou}, \bibinfo{author}{S.~Wu}, \bibinfo{author}{S.~Ye}, \bibinfo{author}{T.~Yun}, \bibinfo{author}{T.~Pei}, \bibinfo{author}{T.~Sun}, \bibinfo{author}{T.~Wang}, \bibinfo{author}{W.~Zeng}, \bibinfo{author}{W.~Zhao}, \bibinfo{author}{W.~Liu}, \bibinfo{author}{W.~Liang}, \bibinfo{author}{W.~Gao}, \bibinfo{author}{W.~Yu}, \bibinfo{author}{W.~Zhang},
  \bibinfo{author}{W.~L. Xiao}, \bibinfo{author}{W.~An}, \bibinfo{author}{X.~Liu}, \bibinfo{author}{X.~Wang}, \bibinfo{author}{X.~Chen}, \bibinfo{author}{X.~Nie}, \bibinfo{author}{X.~Cheng}, \bibinfo{author}{X.~Liu}, \bibinfo{author}{X.~Xie}, \bibinfo{author}{X.~Liu}, \bibinfo{author}{X.~Yang}, \bibinfo{author}{X.~Li}, \bibinfo{author}{X.~Su}, \bibinfo{author}{X.~Lin}, \bibinfo{author}{X.~Q. Li}, \bibinfo{author}{X.~Jin}, \bibinfo{author}{X.~Shen}, \bibinfo{author}{X.~Chen}, \bibinfo{author}{X.~Sun}, \bibinfo{author}{X.~Wang}, \bibinfo{author}{X.~Song}, \bibinfo{author}{X.~Zhou}, \bibinfo{author}{X.~Wang}, \bibinfo{author}{X.~Shan}, \bibinfo{author}{Y.~K. Li}, \bibinfo{author}{Y.~Q. Wang}, \bibinfo{author}{Y.~X. Wei}, \bibinfo{author}{Y.~Zhang}, \bibinfo{author}{Y.~Xu}, \bibinfo{author}{Y.~Li}, \bibinfo{author}{Y.~Zhao}, \bibinfo{author}{Y.~Sun}, \bibinfo{author}{Y.~Wang}, \bibinfo{author}{Y.~Yu}, \bibinfo{author}{Y.~Zhang}, \bibinfo{author}{Y.~Shi}, \bibinfo{author}{Y.~Xiong}, \bibinfo{author}{Y.~He},
  \bibinfo{author}{Y.~Piao}, \bibinfo{author}{Y.~Wang}, \bibinfo{author}{Y.~Tan}, \bibinfo{author}{Y.~Ma}, \bibinfo{author}{Y.~Liu}, \bibinfo{author}{Y.~Guo}, \bibinfo{author}{Y.~Ou}, \bibinfo{author}{Y.~Wang}, \bibinfo{author}{Y.~Gong}, \bibinfo{author}{Y.~Zou}, \bibinfo{author}{Y.~He}, \bibinfo{author}{Y.~Xiong}, \bibinfo{author}{Y.~Luo}, \bibinfo{author}{Y.~You}, \bibinfo{author}{Y.~Liu}, \bibinfo{author}{Y.~Zhou}, \bibinfo{author}{Y.~X. Zhu}, \bibinfo{author}{Y.~Xu}, \bibinfo{author}{Y.~Huang}, \bibinfo{author}{Y.~Li}, \bibinfo{author}{Y.~Zheng}, \bibinfo{author}{Y.~Zhu}, \bibinfo{author}{Y.~Ma}, \bibinfo{author}{Y.~Tang}, \bibinfo{author}{Y.~Zha}, \bibinfo{author}{Y.~Yan}, \bibinfo{author}{Z.~Z. Ren}, \bibinfo{author}{Z.~Ren}, \bibinfo{author}{Z.~Sha}, \bibinfo{author}{Z.~Fu}, \bibinfo{author}{Z.~Xu}, \bibinfo{author}{Z.~Xie}, \bibinfo{author}{Z.~Zhang}, \bibinfo{author}{Z.~Hao}, \bibinfo{author}{Z.~Ma}, \bibinfo{author}{Z.~Yan}, \bibinfo{author}{Z.~Wu}, \bibinfo{author}{Z.~Gu}, \bibinfo{author}{Z.~Zhu},
  \bibinfo{author}{Z.~Liu}, \bibinfo{author}{Z.~Li}, \bibinfo{author}{Z.~Xie}, \bibinfo{author}{Z.~Song}, \bibinfo{author}{Z.~Pan}, \bibinfo{author}{Z.~Huang}, \bibinfo{author}{Z.~Xu}, \bibinfo{author}{Z.~Zhang}, \bibinfo{author}{Z.~Zhang}, \bibinfo{title}{Deepseek-r1: Incentivizing reasoning capability in llms via reinforcement learning}, \bibinfo{year}{2025}. \URLprefix \url{https://arxiv.org/abs/2501.12948}. \href{http://arxiv.org/abs/2501.12948}{{\tt arXiv:2501.12948}}.
\bibitem[{Wei et~al.(2022)Wei, Wang, Schuurmans, Bosma, hsin Chi, Xia, Le, and Zhou}]{wei2022chain}
\bibinfo{author}{J.~Wei}, \bibinfo{author}{X.~Wang}, \bibinfo{author}{D.~Schuurmans}, \bibinfo{author}{M.~Bosma}, \bibinfo{author}{E.~H. hsin Chi}, \bibinfo{author}{F.~Xia}, \bibinfo{author}{Q.~Le}, \bibinfo{author}{D.~Zhou},
\newblock \bibinfo{title}{{Chain of Thought Prompting Elicits Reasoning in Large Language Models}},
\newblock \bibinfo{journal}{ArXiv} \bibinfo{volume}{abs/2201.11903} (\bibinfo{year}{2022}).
\bibitem[{Snell et~al.(2024)Snell, Lee, Xu, and Kumar}]{snell2024scaling}
\bibinfo{author}{C.~Snell}, \bibinfo{author}{J.~Lee}, \bibinfo{author}{K.~Xu}, \bibinfo{author}{A.~Kumar},
\newblock \bibinfo{title}{Scaling llm test-time compute optimally can be more effective than scaling model parameters},
\newblock \bibinfo{journal}{arXiv preprint arXiv:2408.03314}  (\bibinfo{year}{2024}).
\bibitem[{Lambert et~al.(2025)Lambert, Morrison, Pyatkin, Huang, Ivison, Brahman, Miranda, Liu, Dziri, Lyu, Gu, Malik, Graf, Hwang, Yang, Bras, Tafjord, Wilhelm, Soldaini, Smith, Wang, Dasigi, and Hajishirzi}]{lambert2025tulu3pushingfrontiers}
\bibinfo{author}{N.~Lambert}, \bibinfo{author}{J.~Morrison}, \bibinfo{author}{V.~Pyatkin}, \bibinfo{author}{S.~Huang}, \bibinfo{author}{H.~Ivison}, \bibinfo{author}{F.~Brahman}, \bibinfo{author}{L.~J.~V. Miranda}, \bibinfo{author}{A.~Liu}, \bibinfo{author}{N.~Dziri}, \bibinfo{author}{S.~Lyu}, \bibinfo{author}{Y.~Gu}, \bibinfo{author}{S.~Malik}, \bibinfo{author}{V.~Graf}, \bibinfo{author}{J.~D. Hwang}, \bibinfo{author}{J.~Yang}, \bibinfo{author}{R.~L. Bras}, \bibinfo{author}{O.~Tafjord}, \bibinfo{author}{C.~Wilhelm}, \bibinfo{author}{L.~Soldaini}, \bibinfo{author}{N.~A. Smith}, \bibinfo{author}{Y.~Wang}, \bibinfo{author}{P.~Dasigi}, \bibinfo{author}{H.~Hajishirzi}, \bibinfo{title}{{Tulu 3: Pushing Frontiers in Open Language Model Post-Training}}, \bibinfo{year}{2025}. \URLprefix \url{https://arxiv.org/abs/2411.15124}. \href{http://arxiv.org/abs/2411.15124}{{\tt arXiv:2411.15124}}.
\bibitem[{Lewis et~al.(2020)Lewis, Perez, Piktus, Petroni, Karpukhin, Goyal, K{\"u}ttler, Lewis, tau Yih, Rockt{\"a}schel, Riedel, and Kiela}]{lewis2020rag}
\bibinfo{author}{P.~Lewis}, \bibinfo{author}{E.~Perez}, \bibinfo{author}{A.~Piktus}, \bibinfo{author}{F.~Petroni}, \bibinfo{author}{V.~Karpukhin}, \bibinfo{author}{N.~Goyal}, \bibinfo{author}{H.~K{\"u}ttler}, \bibinfo{author}{M.~Lewis}, \bibinfo{author}{W.~tau Yih}, \bibinfo{author}{T.~Rockt{\"a}schel}, \bibinfo{author}{S.~Riedel}, \bibinfo{author}{D.~Kiela},
\newblock \bibinfo{title}{Retrieval-augmented generation for knowledge-intensive {NLP} tasks},
\newblock in: \bibinfo{booktitle}{Advances in Neural Information Processing Systems 33 (NeurIPS 2020)}, \bibinfo{year}{2020}, pp. \bibinfo{pages}{9459--9474}. \URLprefix \url{https://proceedings.neurips.cc/paper/2020/hash/6b493230205f780e1bc26945df7481e5-Abstract.html}.
\bibitem[{Shuster et~al.(2021)Shuster, Poff, Chen, Kiela, and Weston}]{shuster-etal-2021-retrieval-augmentation}
\bibinfo{author}{K.~Shuster}, \bibinfo{author}{S.~Poff}, \bibinfo{author}{M.~Chen}, \bibinfo{author}{D.~Kiela}, \bibinfo{author}{J.~Weston},
\newblock \bibinfo{title}{Retrieval augmentation reduces hallucination in conversation},
\newblock in: \bibinfo{editor}{M.-F. Moens}, \bibinfo{editor}{X.~Huang}, \bibinfo{editor}{L.~Specia}, \bibinfo{editor}{S.~W.-t. Yih} (Eds.), \bibinfo{booktitle}{Findings of the Association for Computational Linguistics: EMNLP 2021}, \bibinfo{publisher}{Association for Computational Linguistics}, \bibinfo{address}{Punta Cana, Dominican Republic}, \bibinfo{year}{2021}, pp. \bibinfo{pages}{3784--3803}. \URLprefix \url{https://aclanthology.org/2021.findings-emnlp.320/}. \DOIprefix\doi{10.18653/v1/2021.findings-emnlp.320}.
\bibitem[{OpenAI(2025)}]{openai2025deepresearch}
\bibinfo{author}{OpenAI}, \bibinfo{title}{{Deep Research System Card}}, \bibinfo{howpublished}{\url{https://cdn.openai.com/deep-research-system-card.pdf}}, \bibinfo{year}{2025}. \bibinfo{note}{System card, accessed 8 Jul 2025}.
\bibitem[{Nakano et~al.(2022)Nakano, Hilton, Balaji, Wu, Ouyang, Kim, Hesse, Jain, Kosaraju, Saunders, Jiang, Cobbe, Eloundou, Krueger, Button, Knight, Chess, and Schulman}]{nakano2022webgptbrowserassistedquestionansweringhuman}
\bibinfo{author}{R.~Nakano}, \bibinfo{author}{J.~Hilton}, \bibinfo{author}{S.~Balaji}, \bibinfo{author}{J.~Wu}, \bibinfo{author}{L.~Ouyang}, \bibinfo{author}{C.~Kim}, \bibinfo{author}{C.~Hesse}, \bibinfo{author}{S.~Jain}, \bibinfo{author}{V.~Kosaraju}, \bibinfo{author}{W.~Saunders}, \bibinfo{author}{X.~Jiang}, \bibinfo{author}{K.~Cobbe}, \bibinfo{author}{T.~Eloundou}, \bibinfo{author}{G.~Krueger}, \bibinfo{author}{K.~Button}, \bibinfo{author}{M.~Knight}, \bibinfo{author}{B.~Chess}, \bibinfo{author}{J.~Schulman}, \bibinfo{title}{Webgpt: Browser-assisted question-answering with human feedback}, \bibinfo{year}{2022}. \URLprefix \url{https://arxiv.org/abs/2112.09332}. \href{http://arxiv.org/abs/2112.09332}{{\tt arXiv:2112.09332}}.
\bibitem[{Russell-Rose et~al.(2021)Russell-Rose, Gooch, and Kruschwitz}]{doi:10.1177/02663821211034079}
\bibinfo{author}{T.~Russell-Rose}, \bibinfo{author}{P.~Gooch}, \bibinfo{author}{U.~Kruschwitz},
\newblock \bibinfo{title}{Interactive query expansion for professional search applications},
\newblock \bibinfo{journal}{Business Information Review} \bibinfo{volume}{38} (\bibinfo{year}{2021}) \bibinfo{pages}{127--137}. \URLprefix \url{https://doi.org/10.1177/02663821211034079}. \DOIprefix\doi{10.1177/02663821211034079}. \href{http://arxiv.org/abs/https://doi.org/10.1177/02663821211034079}{{\tt arXiv:https://doi.org/10.1177/02663821211034079}}.
\bibitem[{Higgins(2008)}]{higgins2008cochrane}
\bibinfo{author}{J.~Higgins},
\newblock \bibinfo{title}{Cochrane handbook for systematic reviews of interventions},
\newblock \bibinfo{journal}{Cochrane Collaboration and John Wiley \& Sons Ltd}  (\bibinfo{year}{2008}).
\bibitem[{Ateia and Kruschwitz(2025)}]{ateia2025bioragent}
\bibinfo{author}{S.~Ateia}, \bibinfo{author}{U.~Kruschwitz},
\newblock \bibinfo{title}{{BioRAGent: A Retrieval-Augmented Generation System for Showcasing Generative Query Expansion and Domain-Specific Search for Scientific Q\&A}},
\newblock in: \bibinfo{booktitle}{European Conference on Information Retrieval}, \bibinfo{year}{2025}.
\bibitem[{Krithara et~al.(2023)Krithara, Nentidis, Bougiatiotis, and Paliouras}]{QAcorpusBioASQ}
\bibinfo{author}{A.~Krithara}, \bibinfo{author}{A.~Nentidis}, \bibinfo{author}{K.~Bougiatiotis}, \bibinfo{author}{G.~Paliouras},
\newblock \bibinfo{title}{{BioASQ-QA: A manually curated corpus for Biomedical Question Answering}},
\newblock \bibinfo{journal}{Scientific Data} \bibinfo{volume}{10} (\bibinfo{year}{2023}) \bibinfo{pages}{170}.
\bibitem[{Ateia(2025)}]{ateia2025professionalsearch}
\bibinfo{author}{S.~Ateia},
\newblock \bibinfo{title}{From professional search to generative deep research systems: How can expert oversight improve search outcomes?},
\newblock in: \bibinfo{booktitle}{Proceedings of the 48th International ACM SIGIR Conference on Research and Development in Information Retrieval (Doctoral Consortium)}, \bibinfo{year}{2025}. \bibinfo{note}{Doctoral Consortium, to appear}.

\end{thebibliography}

\appendix

\section{Detailed Preliminary Results}\label{appendixResults}

\begin{table}[h!]
\caption{Task 13 Phase A Document Retrieval}
\footnotesize
\label{tab:phaseADocumentsBatches}
\begin{tabular}{|c|c|c|c|c|c|c|c|}
\hline
Batch & Position & System & Precision & Recall & F-Measure & \textbf{MAP} & GMAP \\
\hline
Test batch 1 & 1 of 51 & Top Competitor & 0.1047 & 0.5043 & 0.1605 & \textbf{ 0.4246 } & 0.0104 \\
Test batch 1 & \textbf{22} of 51 & UR-IW-5 & 0.1677 & 0.3471 & 0.2038 & \textbf{ 0.2865 } & 0.0015 \\
Test batch 1 & \textbf{26} of 51 & UR-IW-1 & 0.1415 & 0.3194 & 0.1776 & \textbf{ 0.2527 } & 0.0010 \\
Test batch 1 & \textbf{29} of 51 & UR-IW-2 & 0.1376 & 0.2941 & 0.1699 & \textbf{ 0.2272 } & 0.0007 \\
Test batch 1 & \textbf{32} of 51 & UR-IW-3 & 0.1344 & 0.2547 & 0.1557 & \textbf{ 0.2064 } & 0.0005 \\
Test batch 1 & \textbf{34} of 51 & UR-IW-4 & 0.0979 & 0.1892 & 0.1135 & \textbf{ 0.1739 } & 0.0001 \\
\hline
Test batch 2 & 1 of 41 & Top Competitor & 0.0976 & 0.5093 & 0.1546 & \textbf{ 0.4425 } & 0.0096 \\
Test batch 2 & \textbf{25} of 41 & UR-IW-5 & 0.1930 & 0.3237 & 0.2088 & \textbf{ 0.2634 } & 0.0011 \\
Test batch 2 & \textbf{26} of 41 & UR-IW-4 & 0.1575 & 0.3184 & 0.1820 & \textbf{ 0.2601 } & 0.0009 \\
Test batch 2 & \textbf{27} of 41 & UR-IW-1 & 0.1643 & 0.2890 & 0.1855 & \textbf{ 0.2523 } & 0.0008 \\
Test batch 2 & \textbf{28} of 41 & UR-IW-3 & 0.1742 & 0.3064 & 0.1996 & \textbf{ 0.2443 } & 0.0008 \\
Test batch 2 & \textbf{31} of 41 & UR-IW-2 & 0.1181 & 0.2399 & 0.1335 & \textbf{ 0.1846 } & 0.0005 \\
\hline
Test batch 3 & 1 of 47 & Top Competitor & 0.0941 & 0.4228 & 0.1445 & \textbf{ 0.3236 } & 0.0059 \\
Test batch 3 & \textbf{24} of 47 & UR-IW-5 & 0.1341 & 0.2507 & 0.1560 & \textbf{ 0.1834 } & 0.0005 \\
Test batch 3 & \textbf{27} of 47 & UR-IW-1 & 0.1114 & 0.2283 & 0.1273 & \textbf{ 0.1615 } & 0.0004 \\
Test batch 3 & \textbf{30} of 47 & UR-IW-3 & 0.0854 & 0.2086 & 0.1093 & \textbf{ 0.1456 } & 0.0004 \\
Test batch 3 & \textbf{31} of 47 & UR-IW-2 & 0.0703 & 0.1588 & 0.0871 & \textbf{ 0.1187 } & 0.0002 \\
Test batch 3 & \textbf{36} of 47 & UR-IW-4 & 0.0644 & 0.1490 & 0.0818 & \textbf{ 0.1043 } & 0.0001 \\
\hline
Test batch 4 & 1 of 79 & Top Competitor & 0.0600 & 0.2512 & 0.0927 & \textbf{ 0.1801 } & 0.0008 \\
Test batch 4 & \textbf{27} of 79 & UR-IW-5 & 0.0427 & 0.1391 & 0.0632 & \textbf{ 0.0794 } & 0.0002 \\
Test batch 4 & \textbf{31} of 79 & UR-IW-4 & 0.0451 & 0.1371 & 0.0622 & \textbf{ 0.0713 } & 0.0001 \\
Test batch 4 & \textbf{38} of 79 & UR-IW-2 & 0.0408 & 0.1227 & 0.0555 & \textbf{ 0.0655 } & 0.0001 \\
Test batch 4 & \textbf{39} of 79 & UR-IW-1 & 0.0418 & 0.1040 & 0.0522 & \textbf{ 0.0627 } & 0.0001 \\
Test batch 4 & \textbf{45} of 79 & UR-IW-3 & 0.0396 & 0.0900 & 0.0460 & \textbf{ 0.0574 } & 0.0001 \\
\hline
\end{tabular}
\end{table}

\begin{table}[h!]
\caption{Task 13 Phase A Snippet Retrieval}
\footnotesize
\label{tab:phaseASnippetsBatches}
\begin{tabular}{|c|c|c|c|c|c|c|c|}
\hline
Batch & Position & System & Precision & Recall & F-Measure & \textbf{MAP} & GMAP \\
\hline
Test batch 1 & 1 of 51 & Top Competitor & 0.0803 & 0.3050 & 0.1186 & \textbf{ 0.4535 } & 0.0014 \\
Test batch 1 & \textbf{8} of 51 & UR-IW-5 & 0.1189 & 0.1928 & 0.1202 & \textbf{ 0.2768 } & 0.0006 \\
Test batch 1 & \textbf{9} of 51 & UR-IW-1 & 0.0978 & 0.1594 & 0.1071 & \textbf{ 0.2762 } & 0.0005 \\
Test batch 1 & \textbf{10} of 51 & UR-IW-2 & 0.1136 & 0.1633 & 0.1110 & \textbf{ 0.2478 } & 0.0005 \\
Test batch 1 & \textbf{11} of 51 & UR-IW-3 & 0.0863 & 0.1393 & 0.0912 & \textbf{ 0.2447 } & 0.0003 \\
Test batch 1 & \textbf{17} of 51 & UR-IW-4 & 0.0795 & 0.1035 & 0.0778 & \textbf{ 0.1844 } & 0.0001 \\
\hline
Test batch 2 & 1 of 41 & Top Competitor & 0.0941 & 0.3625 & 0.1421 & \textbf{ 0.5522 } & 0.0035 \\
Test batch 2 & \textbf{12} of 41 & UR-IW-5 & 0.1407 & 0.1885 & 0.1290 & \textbf{ 0.3080 } & 0.0009 \\
Test batch 2 & \textbf{13} of 41 & UR-IW-1 & 0.1233 & 0.1713 & 0.1200 & \textbf{ 0.3023 } & 0.0005 \\
Test batch 2 & \textbf{14} of 41 & UR-IW-3 & 0.1287 & 0.1715 & 0.1212 & \textbf{ 0.2949 } & 0.0007 \\
Test batch 2 & \textbf{18} of 41 & UR-IW-4 & 0.1397 & 0.1635 & 0.1149 & \textbf{ 0.2543 } & 0.0007 \\
Test batch 2 & \textbf{24} of 41 & UR-IW-2 & 0.0916 & 0.1287 & 0.0877 & \textbf{ 0.1654 } & 0.0003 \\
\hline
Test batch 3 & 1 of 47 & Top Competitor & 0.0749 & 0.2855 & 0.1098 & \textbf{ 0.4322 } & 0.0015 \\
Test batch 3 & \textbf{15} of 47 & UR-IW-1 & 0.0838 & 0.1160 & 0.0806 & \textbf{ 0.1534 } & 0.0003 \\
Test batch 3 & \textbf{18} of 47 & UR-IW-5 & 0.0961 & 0.1271 & 0.0938 & \textbf{ 0.1488 } & 0.0003 \\
Test batch 3 & \textbf{19} of 47 & UR-IW-3 & 0.0602 & 0.1130 & 0.0745 & \textbf{ 0.1463 } & 0.0002 \\
Test batch 3 & \textbf{24} of 47 & UR-IW-2 & 0.0680 & 0.0662 & 0.0595 & \textbf{ 0.0968 } & 0.0001 \\
Test batch 3 & \textbf{25} of 47 & UR-IW-4 & 0.0478 & 0.0623 & 0.0475 & \textbf{ 0.0721 } & 0.0001 \\
\hline
Test batch 4 & 1 of 79 & Top Competitor & 0.0411 & 0.1135 & 0.0560 & \textbf{ 0.1634 } & 0.0001 \\
Test batch 4 & \textbf{18} of 79 & UR-IW-5 & 0.0308 & 0.0633 & 0.0399 & \textbf{ 0.0511 } & 0.0001 \\
Test batch 4 & \textbf{20} of 79 & UR-IW-3 & 0.0247 & 0.0492 & 0.0297 & \textbf{ 0.0459 } & 0.0000 \\
Test batch 4 & \textbf{21} of 79 & UR-IW-4 & 0.0229 & 0.0564 & 0.0314 & \textbf{ 0.0446 } & 0.0001 \\
Test batch 4 & \textbf{22} of 79 & UR-IW-1 & 0.0254 & 0.0639 & 0.0323 & \textbf{ 0.0434 } & 0.0000 \\
Test batch 4 & \textbf{23} of 79 & UR-IW-2 & 0.0250 & 0.0700 & 0.0344 & \textbf{ 0.0411 } & 0.0001 \\
\hline
\end{tabular}
\end{table}

\begin{table}[h!]
\caption{Task 13 B Phase A+ Yes/No questions}
\footnotesize
\label{tab:13BA+YesNoBatches}
\begin{tabular}{|c|c|c|c|c|c|c|}
\hline
Batch & Position & System & Accuracy & F1 Yes & F1 No & \textbf{Macro F1} \\
\hline
Test batch 1 & 1 of 56 & Top Competitor & 1.0000 & 1.0000 & 1.0000 & \textbf{ 1.0000 } \\
Test batch 1 & \textbf{1} of 56 & UR-IW-1 & 1.0000 & 1.0000 & 1.0000 & \textbf{ 1.0000 } \\
Test batch 1 & \textbf{1} of 56 & UR-IW-2 & 1.0000 & 1.0000 & 1.0000 & \textbf{ 1.0000 } \\
Test batch 1 & \textbf{1} of 56 & UR-IW-4 & 1.0000 & 1.0000 & 1.0000 & \textbf{ 1.0000 } \\
Test batch 1 & \textbf{21} of 56 & UR-IW-3 & 0.9412 & 0.9565 & 0.9091 & \textbf{ 0.9328 } \\
Test batch 1 & \textbf{42} of 56 & UR-IW-5 & 0.8235 & 0.8696 & 0.7273 & \textbf{ 0.7984 } \\
\hline
Test batch 2 & \textbf{1} of 49 & UR-IW-5 & 1.0000 & 1.0000 & 1.0000 & \textbf{ 1.0000 } \\
Test batch 2 & \textbf{10} of 49 & UR-IW-1 & 0.9412 & 0.9565 & 0.9091 & \textbf{ 0.9328 } \\
Test batch 2 & \textbf{13} of 49 & UR-IW-3 & 0.8824 & 0.9000 & 0.8571 & \textbf{ 0.8786 } \\
Test batch 2 & \textbf{14} of 49 & UR-IW-2 & 0.8824 & 0.9091 & 0.8333 & \textbf{ 0.8712 } \\
Test batch 2 & \textbf{37} of 49 & UR-IW-4 & 0.7059 & 0.7368 & 0.6667 & \textbf{ 0.7018 } \\
\hline
Test batch 3 & 1 of 58 & Top Competitor & 0.9545 & 0.9697 & 0.9091 & \textbf{ 0.9394 } \\
Test batch 3 & \textbf{5} of 58 & UR-IW-5 & 0.9091 & 0.9412 & 0.8000 & \textbf{ 0.8706 } \\
Test batch 3 & \textbf{10} of 58 & UR-IW-4 & 0.8636 & 0.9091 & 0.7273 & \textbf{ 0.8182 } \\
Test batch 3 & \textbf{19} of 58 & UR-IW-1 & 0.8636 & 0.9143 & 0.6667 & \textbf{ 0.7905 } \\
Test batch 3 & \textbf{26} of 58 & UR-IW-2 & 0.8182 & 0.8824 & 0.6000 & \textbf{ 0.7412 } \\
Test batch 3 & \textbf{42} of 58 & UR-IW-3 & 0.6818 & 0.7742 & 0.4615 & \textbf{ 0.6179 } \\
\hline
Test batch 4 & 1 of 67 & Top Competitor & 0.9231 & 0.9444 & 0.8750 & \textbf{ 0.9097 } \\
Test batch 4 & \textbf{2} of 67 & UR-IW-2 & 0.9231 & 0.9444 & 0.8750 & \textbf{ 0.9097 } \\
Test batch 4 & \textbf{34} of 67 & UR-IW-3 & 0.8462 & 0.8889 & 0.7500 & \textbf{ 0.8194 } \\
Test batch 4 & \textbf{37} of 67 & UR-IW-1 & 0.8462 & 0.8947 & 0.7143 & \textbf{ 0.8045 } \\
Test batch 4 & \textbf{39} of 67 & UR-IW-4 & 0.8077 & 0.8485 & 0.7368 & \textbf{ 0.7927 } \\
Test batch 4 & \textbf{43} of 67 & UR-IW-5 & 0.8462 & 0.9000 & 0.6667 & \textbf{ 0.7833 } \\
\hline
\end{tabular}
\end{table}

\begin{table}[h!]
\caption{Task 13 B Phase A+ factoid questions}
\footnotesize
\label{tab:13BA+FactoidBatches}
\begin{tabular}{|c|c|c|c|c|c|}
\hline
Batch & Position & System & Strict Acc. & Lenient Acc. & \textbf{MRR} \\
\hline
Test batch 1 & 1 of 56 & Top Competitor & 0.4231 & 0.5000 & \textbf{ 0.4551 } \\
Test batch 1 & \textbf{7} of 56 & UR-IW-2 & 0.3462 & 0.4231 & \textbf{ 0.3782 } \\
Test batch 1 & \textbf{8} of 56 & UR-IW-5 & 0.3462 & 0.4231 & \textbf{ 0.3750 } \\
Test batch 1 & \textbf{24} of 56 & UR-IW-1 & 0.2692 & 0.3462 & \textbf{ 0.3077 } \\
Test batch 1 & \textbf{25} of 56 & UR-IW-3 & 0.2692 & 0.3462 & \textbf{ 0.3077 } \\
Test batch 1 & \textbf{29} of 56 & UR-IW-4 & 0.2692 & 0.3077 & \textbf{ 0.2885 } \\
\hline
Test batch 2 & 1 of 49 & Top Competitor & 0.5926 & 0.5926 & \textbf{ 0.5926 } \\
Test batch 2 & \textbf{2} of 49 & UR-IW-4 & 0.5185 & 0.5556 & \textbf{ 0.5370 } \\
Test batch 2 & \textbf{4} of 49 & UR-IW-2 & 0.4815 & 0.5185 & \textbf{ 0.5000 } \\
Test batch 2 & \textbf{14} of 49 & UR-IW-1 & 0.4074 & 0.4815 & \textbf{ 0.4383 } \\
Test batch 2 & \textbf{16} of 49 & UR-IW-3 & 0.3704 & 0.4444 & \textbf{ 0.3920 } \\
Test batch 2 & \textbf{29} of 49 & UR-IW-5 & 0.2963 & 0.3333 & \textbf{ 0.3086 } \\
\hline
Test batch 3 & 1 of 58 & Top Competitor & 0.3500 & 0.4000 & \textbf{ 0.3750 } \\
Test batch 3 & \textbf{7} of 58 & UR-IW-3 & 0.2000 & 0.5000 & \textbf{ 0.3100 } \\
Test batch 3 & \textbf{12} of 58 & UR-IW-1 & 0.2000 & 0.4000 & \textbf{ 0.2875 } \\
Test batch 3 & \textbf{18} of 58 & UR-IW-2 & 0.2500 & 0.2500 & \textbf{ 0.2500 } \\
Test batch 3 & \textbf{27} of 58 & UR-IW-4 & 0.2000 & 0.2000 & \textbf{ 0.2000 } \\
Test batch 3 & \textbf{29} of 58 & UR-IW-5 & 0.1000 & 0.3000 & \textbf{ 0.2000 } \\
\hline
Test batch 4 & \textbf{1} of 67 & UR-IW-5 & 0.5455 & 0.5909 & \textbf{ 0.5606 } \\
Test batch 4 & \textbf{4} of 67 & UR-IW-1 & 0.5000 & 0.5455 & \textbf{ 0.5152 } \\
Test batch 4 & \textbf{13} of 67 & UR-IW-2 & 0.4545 & 0.4545 & \textbf{ 0.4545 } \\
Test batch 4 & \textbf{14} of 67 & UR-IW-3 & 0.4545 & 0.4545 & \textbf{ 0.4545 } \\
Test batch 4 & \textbf{34} of 67 & UR-IW-4 & 0.3636 & 0.4091 & \textbf{ 0.3788 } \\
\hline
\end{tabular}
\end{table}

\begin{table}[h!]
\caption{Task 13 B Phase A+ list questions}
\footnotesize
\label{tab:13BA+ListBatches}
\begin{tabular}{|c|c|c|c|c|c|}
\hline
Batch & Position & System & Mean Prec. & Recall & \textbf{F-Measure} \\
\hline
Test batch 1 & \textbf{1} of 56 & UR-IW-2 & 0.2290 & 0.3056 & \textbf{ 0.2567 } \\
Test batch 1 & \textbf{2} of 56 & UR-IW-1 & 0.2070 & 0.3232 & \textbf{ 0.2411 } \\
Test batch 1 & \textbf{3} of 56 & UR-IW-5 & 0.2164 & 0.3003 & \textbf{ 0.2395 } \\
Test batch 1 & \textbf{4} of 56 & UR-IW-4 & 0.2134 & 0.2783 & \textbf{ 0.2357 } \\
Test batch 1 & \textbf{15} of 56 & UR-IW-3 & 0.1685 & 0.2804 & \textbf{ 0.2004 } \\
\hline
Test batch 2 & 1 of 49 & Top Competitor & 0.3785 & 0.4357 & \textbf{ 0.3880 } \\
Test batch 2 & \textbf{2} of 49 & UR-IW-2 & 0.3449 & 0.4626 & \textbf{ 0.3805 } \\
Test batch 2 & \textbf{11} of 49 & UR-IW-4 & 0.2859 & 0.3652 & \textbf{ 0.3023 } \\
Test batch 2 & \textbf{17} of 49 & UR-IW-1 & 0.2307 & 0.3536 & \textbf{ 0.2682 } \\
Test batch 2 & \textbf{33} of 49 & UR-IW-5 & 0.1796 & 0.3432 & \textbf{ 0.2144 } \\
Test batch 2 & \textbf{34} of 49 & UR-IW-3 & 0.1696 & 0.3213 & \textbf{ 0.2118 } \\
\hline
Test batch 3 & 1 of 58 & Top Competitor & 0.4674 & 0.4446 & \textbf{ 0.4541 } \\
Test batch 3 & \textbf{13} of 58 & UR-IW-5 & 0.3455 & 0.4111 & \textbf{ 0.3618 } \\
Test batch 3 & \textbf{14} of 58 & UR-IW-2 & 0.3656 & 0.3969 & \textbf{ 0.3599 } \\
Test batch 3 & \textbf{20} of 58 & UR-IW-1 & 0.3271 & 0.4040 & \textbf{ 0.3482 } \\
Test batch 3 & \textbf{26} of 58 & UR-IW-3 & 0.3114 & 0.3777 & \textbf{ 0.3279 } \\
Test batch 3 & \textbf{37} of 58 & UR-IW-4 & 0.2206 & 0.2896 & \textbf{ 0.2371 } \\
\hline
Test batch 4 & 1 of 67 & Top Competitor & 0.3217 & 0.2929 & \textbf{ 0.3014 } \\
Test batch 4 & \textbf{9} of 67 & UR-IW-5 & 0.2345 & 0.3680 & \textbf{ 0.2742 } \\
Test batch 4 & \textbf{11} of 67 & UR-IW-3 & 0.2640 & 0.3114 & \textbf{ 0.2739 } \\
Test batch 4 & \textbf{34} of 67 & UR-IW-4 & 0.2122 & 0.2936 & \textbf{ 0.2270 } \\
Test batch 4 & \textbf{37} of 67 & UR-IW-1 & 0.1819 & 0.3429 & \textbf{ 0.2246 } \\
Test batch 4 & \textbf{42} of 67 & UR-IW-2 & 0.1846 & 0.3349 & \textbf{ 0.2172 } \\
\hline
\end{tabular}
\end{table}

\begin{table}[h!]
\caption{Task 13 B Phase B Yes/No questions}
\footnotesize
\label{tab:task13BYesNoBatches}
\begin{tabular}{|c|c|c|c|c|c|c|}
\hline
Batch & Position & System & Accuracy & F1 Yes & F1 No & \textbf{Macro F1} \\
\hline
Test batch 1 & 1 of 72 & Top Competitor & 1.0000 & 1.0000 & 1.0000 & \textbf{ 1.0000 } \\
Test batch 1 & \textbf{1} of 72 & UR-IW-1 & 1.0000 & 1.0000 & 1.0000 & \textbf{ 1.0000 } \\
Test batch 1 & \textbf{1} of 72 & UR-IW-5 & 1.0000 & 1.0000 & 1.0000 & \textbf{ 1.0000 } \\
Test batch 1 & \textbf{36} of 72 & UR-IW-2 & 0.9412 & 0.9600 & 0.8889 & \textbf{ 0.9244 } \\
Test batch 1 & \textbf{53} of 72 & UR-IW-3 & 0.8824 & 0.9091 & 0.8333 & \textbf{ 0.8712 } \\
Test batch 1 & \textbf{60} of 72 & UR-IW-4 & 0.8235 & 0.8696 & 0.7273 & \textbf{ 0.7984 } \\
\hline
Test batch 2 & \textbf{1} of 72 & UR-IW-5 & 1.0000 & 1.0000 & 1.0000 & \textbf{ 1.0000 } \\
Test batch 2 & \textbf{35} of 72 & UR-IW-4 & 0.9412 & 0.9565 & 0.9091 & \textbf{ 0.9328 } \\
Test batch 2 & \textbf{47} of 72 & UR-IW-3 & 0.8824 & 0.9000 & 0.8571 & \textbf{ 0.8786 } \\
Test batch 2 & \textbf{52} of 72 & UR-IW-2 & 0.8824 & 0.9091 & 0.8333 & \textbf{ 0.8712 } \\
Test batch 2 & \textbf{56} of 72 & UR-IW-1 & 0.8824 & 0.9167 & 0.8000 & \textbf{ 0.8583 } \\
\hline
Test batch 3 & 1 of 66 & Top Competitor & 0.9545 & 0.9697 & 0.9091 & \textbf{ 0.9394 } \\
Test batch 3 & \textbf{30} of 66 & UR-IW-4 & 0.9091 & 0.9412 & 0.8000 & \textbf{ 0.8706 } \\
Test batch 3 & \textbf{31} of 66 & UR-IW-5 & 0.9091 & 0.9412 & 0.8000 & \textbf{ 0.8706 } \\
Test batch 3 & \textbf{42} of 66 & UR-IW-2 & 0.8636 & 0.9091 & 0.7273 & \textbf{ 0.8182 } \\
Test batch 3 & \textbf{43} of 66 & UR-IW-3 & 0.8636 & 0.9091 & 0.7273 & \textbf{ 0.8182 } \\
Test batch 3 & \textbf{52} of 66 & UR-IW-1 & 0.8636 & 0.9143 & 0.6667 & \textbf{ 0.7905 } \\
\hline
Test batch 4 & 1 of 79 & Top Competitor & 1.0000 & 1.0000 & 1.0000 & \textbf{ 1.0000 } \\
Test batch 4 & \textbf{21} of 79 & UR-IW-4 & 0.9231 & 0.9444 & 0.8750 & \textbf{ 0.9097 } \\
Test batch 4 & \textbf{30} of 79 & UR-IW-2 & 0.9231 & 0.9474 & 0.8571 & \textbf{ 0.9023 } \\
Test batch 4 & \textbf{31} of 79 & UR-IW-5 & 0.9231 & 0.9474 & 0.8571 & \textbf{ 0.9023 } \\
Test batch 4 & \textbf{45} of 79 & UR-IW-1 & 0.9231 & 0.9500 & 0.8333 & \textbf{ 0.8917 } \\
Test batch 4 & \textbf{66} of 79 & UR-IW-3 & 0.7692 & 0.8235 & 0.6667 & \textbf{ 0.7451 } \\
\hline
\end{tabular}
\end{table}

\begin{table}[h!]
\caption{Task 13 B Phase B factoid questions}
\footnotesize
\label{tab:task13BFactoidBatches}
\begin{tabular}{|c|c|c|c|c|c|}
\hline
Batch & Position & System & Strict Acc. & Lenient Acc. & \textbf{MRR} \\
\hline
Test batch 1 & 1 of 72 & Top Competitor & 0.5385 & 0.6538 & \textbf{ 0.5962 } \\
Test batch 1 & \textbf{17} of 72 & UR-IW-3 & 0.4231 & 0.5769 & \textbf{ 0.4821 } \\
Test batch 1 & \textbf{26} of 72 & UR-IW-4 & 0.4231 & 0.5000 & \textbf{ 0.4615 } \\
Test batch 1 & \textbf{27} of 72 & UR-IW-5 & 0.4231 & 0.5000 & \textbf{ 0.4615 } \\
Test batch 1 & \textbf{33} of 72 & UR-IW-2 & 0.4231 & 0.5000 & \textbf{ 0.4551 } \\
Test batch 1 & \textbf{37} of 72 & UR-IW-1 & 0.3846 & 0.5385 & \textbf{ 0.4423 } \\
\hline
Test batch 2 & 1 of 72 & Top Competitor & 0.7037 & 0.7037 & \textbf{ 0.7037 } \\
Test batch 2 & \textbf{11} of 72 & UR-IW-1 & 0.5556 & 0.6296 & \textbf{ 0.5926 } \\
Test batch 2 & \textbf{21} of 72 & UR-IW-2 & 0.5185 & 0.5926 & \textbf{ 0.5556 } \\
Test batch 2 & \textbf{22} of 72 & UR-IW-4 & 0.5185 & 0.5926 & \textbf{ 0.5556 } \\
Test batch 2 & \textbf{30} of 72 & UR-IW-5 & 0.5185 & 0.5556 & \textbf{ 0.5370 } \\
Test batch 2 & \textbf{35} of 72 & UR-IW-3 & 0.5185 & 0.5556 & \textbf{ 0.5309 } \\
\hline
Test batch 3 & 1 of 66 & Top Competitor & 0.4000 & 0.6500 & \textbf{ 0.5100 } \\
Test batch 3 & \textbf{24} of 66 & UR-IW-1 & 0.3500 & 0.4500 & \textbf{ 0.3725 } \\
Test batch 3 & \textbf{31} of 66 & UR-IW-4 & 0.3000 & 0.3500 & \textbf{ 0.3250 } \\
Test batch 3 & \textbf{32} of 66 & UR-IW-5 & 0.2500 & 0.4000 & \textbf{ 0.3250 } \\
Test batch 3 & \textbf{38} of 66 & UR-IW-3 & 0.2500 & 0.3500 & \textbf{ 0.3000 } \\
Test batch 3 & \textbf{43} of 66 & UR-IW-2 & 0.2500 & 0.3000 & \textbf{ 0.2750 } \\
\hline
Test batch 4 & 1 of 79 & Top Competitor & 0.6364 & 0.6364 & \textbf{ 0.6364 } \\
Test batch 4 & \textbf{6} of 79 & UR-IW-4 & 0.5455 & 0.6364 & \textbf{ 0.5909 } \\
Test batch 4 & \textbf{17} of 79 & UR-IW-1 & 0.5455 & 0.5909 & \textbf{ 0.5606 } \\
Test batch 4 & \textbf{27} of 79 & UR-IW-5 & 0.5000 & 0.5455 & \textbf{ 0.5227 } \\
Test batch 4 & \textbf{30} of 79 & UR-IW-2 & 0.5000 & 0.5000 & \textbf{ 0.5000 } \\
Test batch 4 & \textbf{46} of 79 & UR-IW-3 & 0.4545 & 0.4545 & \textbf{ 0.4545 } \\
\hline
\end{tabular}
\end{table}

\begin{table}[h!]
\caption{Task 13 B Phase B List questions}
\footnotesize
\label{tab:task13BListBatches}
\begin{tabular}{|c|c|c|c|c|c|}
\hline
Batch & Position & System & Mean Prec. & Recall & \textbf{F-Measure} \\
\hline
Test batch 1 & 1 of 72 & Top Competitor & 0.5820 & 0.6224 & \textbf{ 0.5959 } \\
Test batch 1 & \textbf{28} of 72 & UR-IW-4 & 0.4817 & 0.5601 & \textbf{ 0.5069 } \\
Test batch 1 & \textbf{50} of 72 & UR-IW-2 & 0.3740 & 0.4944 & \textbf{ 0.4042 } \\
Test batch 1 & \textbf{52} of 72 & UR-IW-1 & 0.3361 & 0.5653 & \textbf{ 0.3978 } \\
Test batch 1 & \textbf{57} of 72 & UR-IW-3 & 0.3199 & 0.5419 & \textbf{ 0.3769 } \\
Test batch 1 & \textbf{60} of 72 & UR-IW-5 & 0.2877 & 0.5341 & \textbf{ 0.3515 } \\
\hline
Test batch 2 & 1 of 72 & Top Competitor & 0.6360 & 0.6315 & \textbf{ 0.6152 } \\
Test batch 2 & \textbf{28} of 72 & UR-IW-4 & 0.4610 & 0.6425 & \textbf{ 0.5188 } \\
Test batch 2 & \textbf{31} of 72 & UR-IW-3 & 0.4312 & 0.6771 & \textbf{ 0.5010 } \\
Test batch 2 & \textbf{41} of 72 & UR-IW-1 & 0.4088 & 0.6561 & \textbf{ 0.4716 } \\
Test batch 2 & \textbf{47} of 72 & UR-IW-5 & 0.3916 & 0.6048 & \textbf{ 0.4463 } \\
Test batch 2 & \textbf{49} of 72 & UR-IW-2 & 0.3833 & 0.5784 & \textbf{ 0.4407 } \\
\hline
Test batch 3 & 1 of 66 & Top Competitor & 0.6433 & 0.6429 & \textbf{ 0.6337 } \\
Test batch 3 & \textbf{44} of 66 & UR-IW-5 & 0.4465 & 0.5790 & \textbf{ 0.4783 } \\
Test batch 3 & \textbf{45} of 66 & UR-IW-3 & 0.4324 & 0.6102 & \textbf{ 0.4774 } \\
Test batch 3 & \textbf{46} of 66 & UR-IW-4 & 0.4472 & 0.5272 & \textbf{ 0.4687 } \\
Test batch 3 & \textbf{47} of 66 & UR-IW-1 & 0.4371 & 0.6368 & \textbf{ 0.4684 } \\
Test batch 3 & \textbf{54} of 66 & UR-IW-2 & 0.4009 & 0.5549 & \textbf{ 0.4369 } \\
\hline
Test batch 4 & 1 of 79 & Top Competitor & 0.7491 & 0.5980 & \textbf{ 0.6492 } \\
Test batch 4 & \textbf{41} of 79 & UR-IW-3 & 0.4140 & 0.6127 & \textbf{ 0.4711 } \\
Test batch 4 & \textbf{46} of 79 & UR-IW-4 & 0.4163 & 0.5536 & \textbf{ 0.4479 } \\
Test batch 4 & \textbf{47} of 79 & UR-IW-5 & 0.3671 & 0.5911 & \textbf{ 0.4338 } \\
Test batch 4 & \textbf{55} of 79 & UR-IW-1 & 0.3303 & 0.5425 & \textbf{ 0.3872 } \\
Test batch 4 & \textbf{56} of 79 & UR-IW-2 & 0.3285 & 0.5413 & \textbf{ 0.3797 } \\
\hline
\end{tabular}
\end{table}

\end{document}